\documentclass[10pt,twocolumn,letterpaper]{article}

\usepackage{iccv}
\usepackage{times}
\usepackage{epsfig}
\usepackage{graphicx}
\usepackage{amsmath}
\usepackage{amssymb}

\usepackage{caption}
\usepackage{subcaption}
\usepackage{tabulary}
\usepackage{listings}
\usepackage{bbm}

\usepackage[breaklinks=true,bookmarks=false]{hyperref}

\iccvfinalcopy 

\def\iccvPaperID{3034} 

\ificcvfinal\fi

\begin{document}

\title{Topological Map Extraction From Overhead Images}

\author{Zuoyue Li\textsuperscript{1}, Jan Dirk Wegner\textsuperscript{2}, Aur\'elien Lucchi\textsuperscript{1}\\
ETH Z\"urich, Switzerland\\
{\tt\small \textsuperscript{\textnormal{1}}\{li.zuoyue,aurelien.lucchi\}@inf.ethz.ch, \textsuperscript{\textnormal{2}}jan.wegner@geod.baug.ethz.ch}
}

\maketitle
\ificcvfinal\thispagestyle{empty}\fi

\begin{abstract}
   \vspace{-0.5em}
   We propose a new approach, named PolyMapper, to circumvent the conventional pixel-wise segmentation of (aerial) images and predict objects in a vector representation directly. PolyMapper directly extracts the topological map of a city from overhead images as collections of building footprints and road networks. In order to unify the shape representation for different types of objects, we also propose a novel sequentialization method that reformulates a graph structure as closed polygons. Experiments are conducted on both existing and self-collected large-scale datasets of several cities. Our empirical results demonstrate that our end-to-end learnable model is capable of drawing polygons of building footprints and road networks that very closely approximate the structure of existing online map services, in a fully automated manner. Quantitative and qualitative comparison to the state-of-the-art also shows that our approach achieves good levels of performance. To the best of our knowledge, the automatic extraction of large-scale topological maps is a novel contribution in the remote sensing community that we believe will help develop models with more informed geometrical constraints.
   \vspace{-0.5em}
\end{abstract}


\section{Introduction}\label{sec:intro}

A fundamental research task in computer vision is pixel-accurate image segmentation, where steady progress has been measured with benchmark challenges such as~\cite{mscoco,everingham2010pascal,everingham2015}. The classical approach in this field consists of assigning a label to each image pixel describing what category it belongs to, thus yielding a labeled image as output. However, for many applications, this is not the final desired output from a user's point of view. In this paper, we will instead focus on applications that require a graph or polygon representation as output. Our interest will be in developing a method that, from an input image, directly produces a polygon representation that describes geometric objects using a vector data structure. Motivated by the success of recent works~\cite{duan2015image, castrejon2017annotating, bauchet2018kippi, acuna2018efficient}, we avoid explicit pixel-wise labeling altogether, but instead directly predict polygons from images in an end-to-end learnable approach.

\begin{figure}[!ht]
    \centering
    \includegraphics[width=\columnwidth]{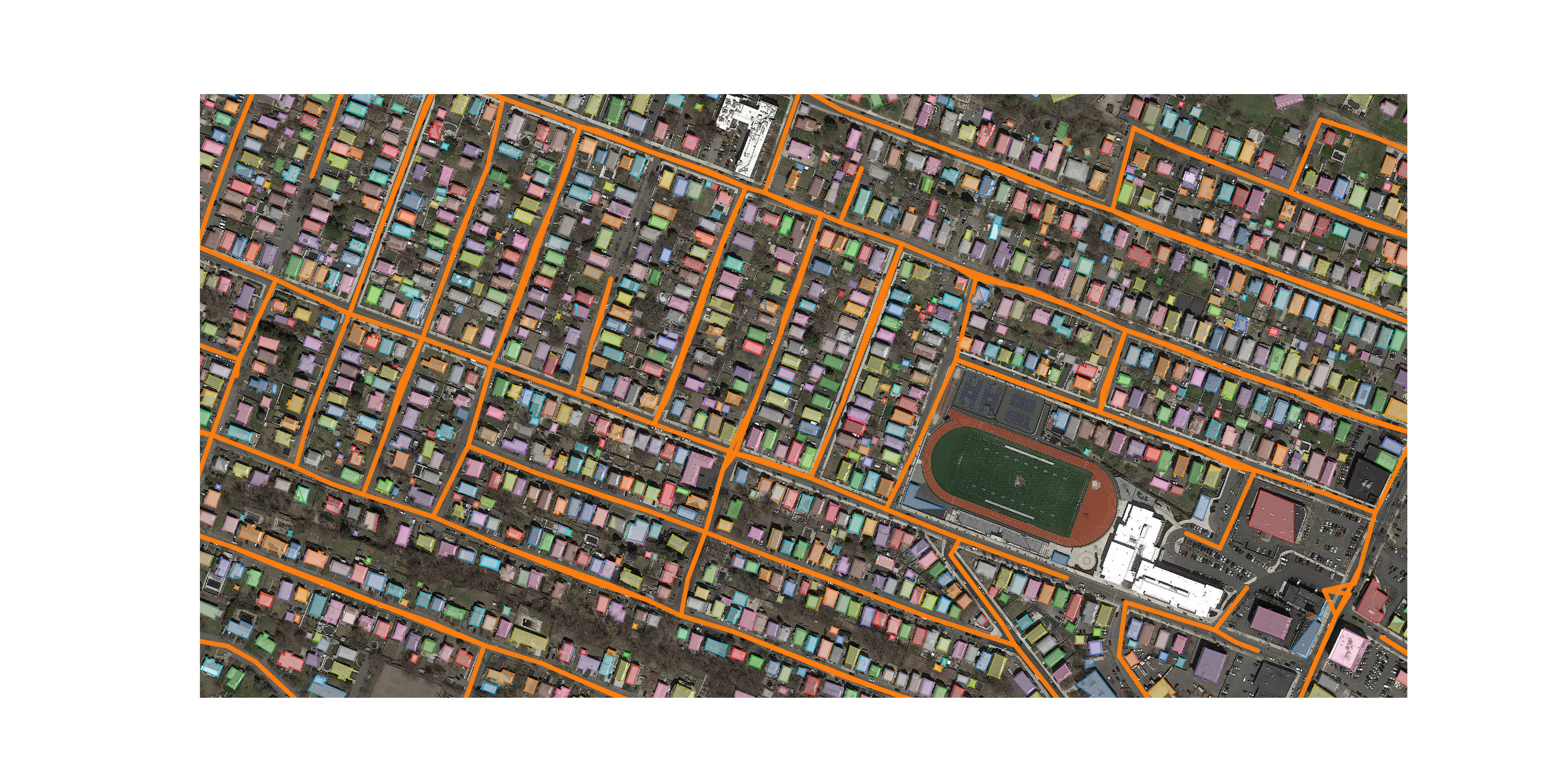}
    \caption{PolyMapper result for Boston overlaid on top of the original aerial imagery. Buildings and roads are directly predicted as polygons. See additional results in Fig.~\ref{fig:other_demos}.}
    \label{fig:demo_boston}
    \vspace{-1.5em}
\end{figure}

Our research is inspired by the insight that for many applications, image segmentation is just an intermediate step of a more comprehensive workflow that aims at a higher-level, abstract, vectorized representation of the image content. A good example is automated map generation from aerial imagery where existing research has mostly focused on aerial image segmentation such as~\cite{dallamura2010,tokarczyk2013,volpi2015,marmanis2016,kaiser2017learning,volpi2017,marmanis2018}. We make this application our core scenario because we have access to virtually unlimited data from OpenStreetMap (OSM)~\cite{haklay2008,haklay2010,girres2010} and high-resolution RGB orthophotos from Google Maps.

Usually, a full mapping pipeline consists of converting an orthophoto to a semantically meaningful raster map (\textit{i.e.}, semantic segmentation), followed by further processing such as object shape refinement, vectorization, and map generalization techniques. Here, we turn this multi-step workflow into an end-to-end learnable deep learning architecture, PolyMapper, which outputs topological maps of buildings and roads directly, given aerial imagery as input.

Our approach performs object detection, instance segmentation, and vectorization within a unified approach that relies on modern CNNs architectures and RNNs with convolutional long-short term memory (ConvLSTM)~\cite{convlstm} modules. As illustrated in Fig.~\ref{fig:rnn}, the CNN takes as input a city tile and extracts keypoints and edge evidence of building footprints and road networks, which are fed sequentially to the multi-layer ConvLSTM modules. The latter produces a vector representation for each object in a given tile. In the case of roads, we also propose an approach that reformulates the topology of roads (typically an undirected graph) as polygons by following a maze solving algorithm that guarantees the shape consistency (sequences) of different objects (see Sec.~\ref{sec:roads}). Finally, the roads from different tiles are connected and combined with the buildings to form a complete city map. A PolyMapper result for the city Boston is shown in Fig.~\ref{fig:demo_boston}, while the results of Chicago and Sunnyvale are illustrated in Fig.~\ref{fig:other_demos}.

We validate our approach for the automated mapping of road networks and building footprints on the existing publicly available datasets and the new collected PolyMapper dataset. Experiment results (see Sec.~\ref{sec:exp}) outperform or are are on par with the state-of-the-art, per-pixel instance segmentation methods~\cite{he2017mask,liu2018path}, and recent research that proposes custom-tailored approaches for only one of the tasks, road network prediction~\cite{deeproadmapper,bastani2018roadtracer} or building footprint extraction~\cite{crowdAIMappingChallengeBaseline2018}. Our approach has significant advantage that it generalizes to both, building and road delineation, and could potentially be extended to other objects.

\section{Related work}\label{RelatedWork}

\begin{figure*}[!ht]
	\centering
    \includegraphics[width=0.9\textwidth]{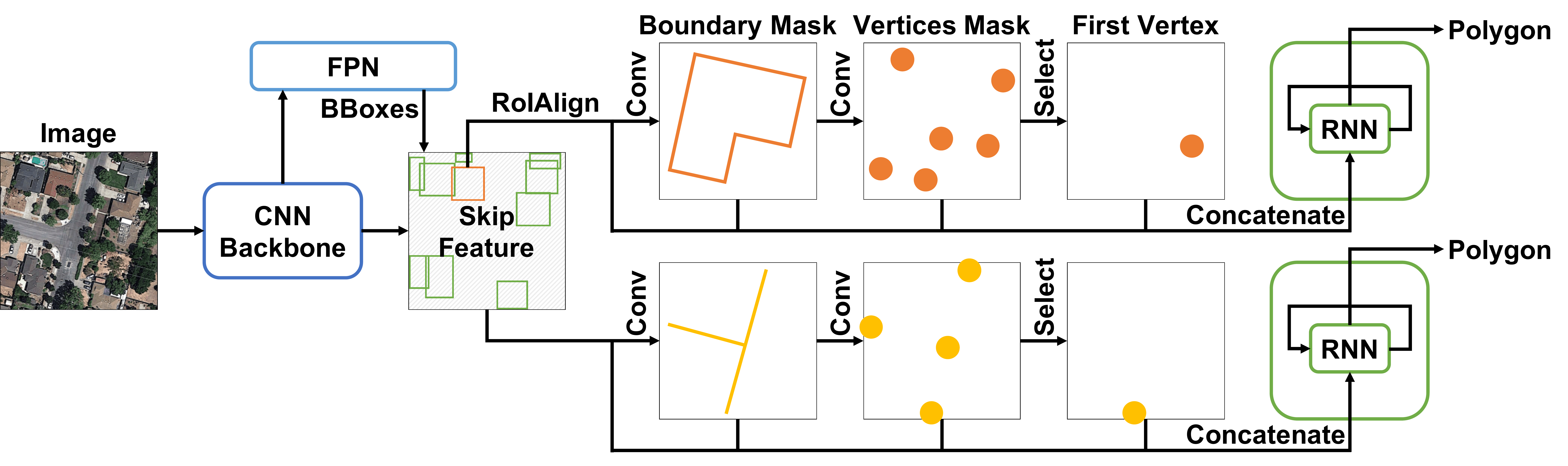}
    \caption{Workflow of our method for both building footprint and road network extraction. The only difference between road and building processing is that we use the corresponding local skip feature via RoIAlign for buildings (bounding boxes provided by FPN), but the entire feature map for roads.}
    \label{fig:cnn_building_road}
    \vspace{-1.5em}
\end{figure*}

\textbf{Building segmentation} from overhead data has been a core research interest for decades and discussing all works is beyond the scope of this paper~\cite{heipke1997,mayer2006,kaiser2017learning}. Before the comeback of deep learning, building footprints were often delineated with multi-step, bottom-up approaches and a combination of multi-spectral overhead imagery and airborne LiDAR, e.g.,~\cite{sohn2007,awrangjeb2010}. A modern approach is~\cite{bittner2018} that applies a fully convolutional neural network to combine evidence from optical overhead imagery and a digital surface model to jointly reason about building footprints. Today, most building footprint delineation from a single image is often approached via semantic segmentation as part of a broader multi-class task and many works exist, e.g.,~\cite{paisitkriangkrai2015,lagrange2015benchmarking,marmanis2016,volpi2017,kaiser2017learning,marmanis2018}.  Microsoft recently extracted all building footprints in the US from aerial images by, first, running semantic segmentation with a CNN and second, refining footprints with a heuristic polygonization approach\footnote{We are not aware of any scientific publication of this work and thus refer the reader to the corresponding \href{https://github.com/Microsoft/USBuildingFootprints}{GitHub repository} that describes the workflow and shares data.}. A current benchmark challenge that aims at extracting building footprints is~\cite{crowdAIMappingChallengeBaseline2018}, which we use to evaluate performance of our approach. Another large-scale dataset that includes both, building footprints and road networks is SpaceNet~\cite{vanetten2018}. All processing takes place in the Amazon Cloud on satellite images of lower resolution than our aerial images in this paper.

\textbf{Road network extraction} in images goes back to (at least) \cite{bajcsy1976}, where road pixels were identified using several image processing operations at a local scale. Shortly afterwards \cite{fischler1981} was probably the first work to explicitly incorporate topology, by searching for long 1-dimensional structures. One of the most sophisticated methods of the pre-deep learning era was introduced in~\cite{stoica2004,lacoste2005}, who center their approach on marked point processes (MPP) that allows them to include elaborate priors on the connectivity and intersection geometry of roads. To the best of our knowledge, the first (non-convolutional) deep learning approach to road network extraction was proposed by~\cite{Mnih2010,mnih2012}. The authors train deep belief network to detect image patches containing roads and second network repairs small network gaps at large scale. ~\cite{wegner13road} propose to model longevity and connectivity of road networks with a higher-order CRF, which is extended in~\cite{wegner2015} to sampling more flexible, road-like higher-order cliques through collections of shortest paths, and to also model buildings with higher-order cliques in \cite{montoya2015}.~\cite{mattyus2015} combine OSM and aerial images to augment maps with additional information like the road width using a MRF formulation, which scales to large regions and achieves good results at several locations world-wide. Two recent works apply deep learning to road center-line extraction in aerial images. DeepRoadMapper~\cite{deeproadmapper} introduces a hierarchical processing pipeline that first segments roads with CNNs, encodes end points of street segments as vertices in a graph connected with edges, thins output segments to road center-lines and repairs gaps with an augmented road graph. RoadTracer~\cite{bastani2018roadtracer} uses an iterative search process guided by a CNN-based decision function to derive the road  network  graph  directly  from  the  output  of  the CNN. To the best of our knowledge,~\cite{bastani2018roadtracer} is the only work, yet, that completely eliminates the intermediate, explicit pixel-wise image labeling step and outputs road center-lines directly like our method.

\textbf{Polygon prediction} in images has a long history with methods such as level sets~\cite{sethian1996} or active contour models~\cite{kass1988}. While these methods follow an iterative energy minimization scheme and usually are a final component of multi-step, bottom-up workflows (e.g.,~\cite{butenuth2012,goepfert2012} for road network refinement), directly predicting polygons from images is a relatively new research direction. We are aware of only six works that move away from pixel-wise labeling and directly predict 2D polygons~\cite{duan2015image, castrejon2017annotating,bastani2018roadtracer,bauchet2018kippi, acuna2018efficient,marcos2018}. Interestingly,~\cite{duan2015image,bauchet2018kippi} apply an unsupervised strategy without making use of deep learning and achieve good results for super-pixel polygons~\cite{duan2015image} and polygonal object segmentation~\cite{bauchet2018kippi}. ~\cite{castrejon2017annotating} designed a semi-automated approach where a human annotator first provide bounding boxes surrounding an object of interest. A deep-learning approach consisting of an RNN coupled with a CNN, then generates a polygon outlining the target object. A recent extension of this work~\cite{acuna2018efficient} increases the output resolution by adding a graph neural network (GNN)~\cite{gnn, li15gated}. This approach, as well as the original work of~\cite{castrejon2017annotating}, still relies on user input to provide an initial bounding box around the object of interest, or to correct a predicted vertex of the polygon if needed. \cite{marcos2018} extracts building footprints by formulating active contours as a deep learning task, where a structured loss imposes learned shape priors that refine an initial extraction result.

In summary, prior works mentioned above either focus on pixel level outputs or can only handle just a single type of object. Thus, the absence of direct topological map extraction in the field of remote sensing is what motivates us to develop a fully automated, end-to-end learnable approach to detect geometrical shapes of buildings and roads in a given overhead image.
\section{Method}
\label{sec:method}

We introduce a new, generic approach for extracting topological map in aerial images using polygons. We first start by discussing the use of polygon representations to describe objects in an image.

\subsection{Polygon Representation}
\label{sec:polygon_representation}

We represent objects as polygons. As in~\cite{castrejon2017annotating, acuna2018efficient}, we rely on a CNN to find keypoints based on image evidence, which are then connected sequentially by an RNN. A fundamental difference of PolyMapper is that it runs fully automatically without any human intervention in contrast to~\cite{castrejon2017annotating, acuna2018efficient}, which were originally designed for speeding up manual object annotation. All the models discussed in~\cite{castrejon2017annotating, acuna2018efficient} (including their ``prediction mode'') require a user to first draw a bounding box that contains the target object and potentially provide additional manual intervention (\textit{e.g.}, drag/add/delete some keypoints) if the object is not correctly delineated.

We refrain from any manual intervention altogether and propose a fully automated workflow. This is however difficult for mainly two reasons: (1) multiple objects of interest can appear in a given image patch and (2) the shapes of different target objects can significantly vary. For instance, buildings are closed shapes of limited extent in the image while road networks span across entire scenes and are best described with a general graph topology. We therefore present two enhancements to address these problems and then introduce the general pipeline as shown in Fig.~\ref{fig:cnn_building_road} for generating object polygons.

\subsection{Multiple Targets}
\label{sec:buildings}

Prior work such as~\cite{castrejon2017annotating, acuna2018efficient} is only applicable when a bounding box is provided for each object of interest. These methods are therefore not able to detect objects such as multiple buildings in a given image. We first address the case of buildings by adding a bounding box detection step to partition the image into individual building instances, which allows to compute separate polygons for all buildings. To this end, we have integrated the Feature Pyramid Network (FPN)~\cite{lin17fpn} into our workflow and have made it an end-to-end model. The FPN further enhances the performance of the region proposal network (RPN) used by Faster R-CNN~\cite{ren2015faster} by exploiting the multi-scale, pyramidal hierarchy of CNNs and resulting in a set of so-called feature pyramids. Once images with individual buildings have been generated, the rest of the pipeline follows the generic procedure described in Sec.~\ref{sec:pipeline}.

\subsection{From Graphs to Polygons}
\label{sec:roads}
\begin{figure}[!ht]
	\centering
	\begin{subfigure}{0.3269\columnwidth}
	    \centering
        \includegraphics[width=\textwidth]{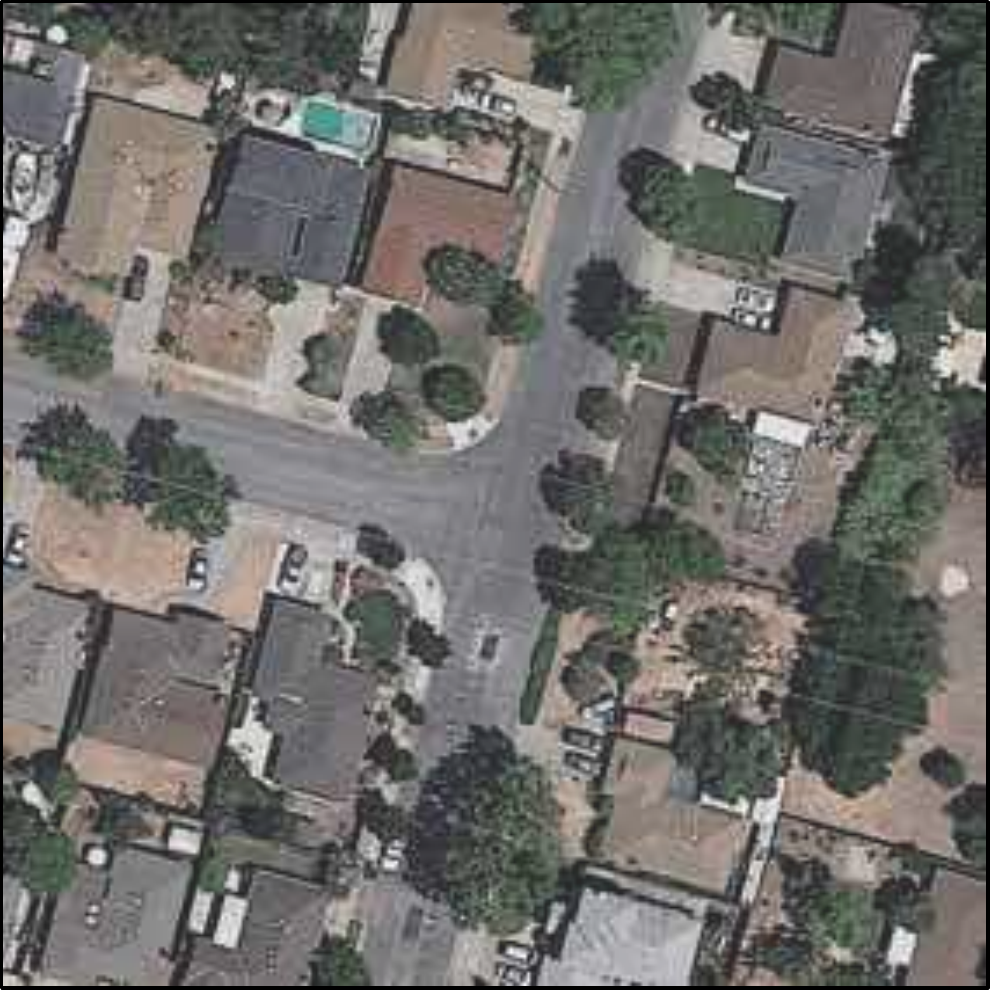}
        \caption{}
        \label{fig:path_img}
    \end{subfigure}
    \begin{subfigure}{0.3269\columnwidth}
        \centering
        \includegraphics[width=\textwidth]{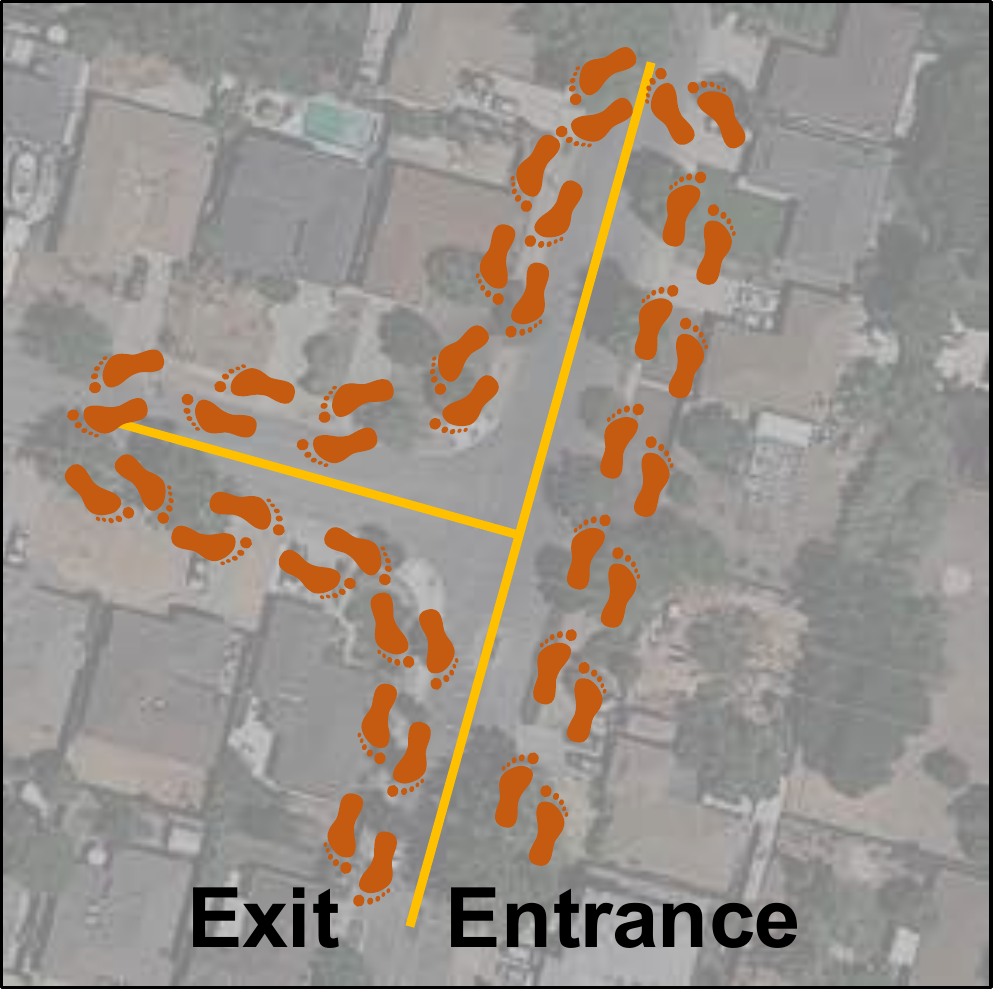}
        \caption{}
        \label{fig:path_true}
    \end{subfigure}
    \begin{subfigure}{0.3269\columnwidth}
        \centering
        \includegraphics[width=\textwidth]{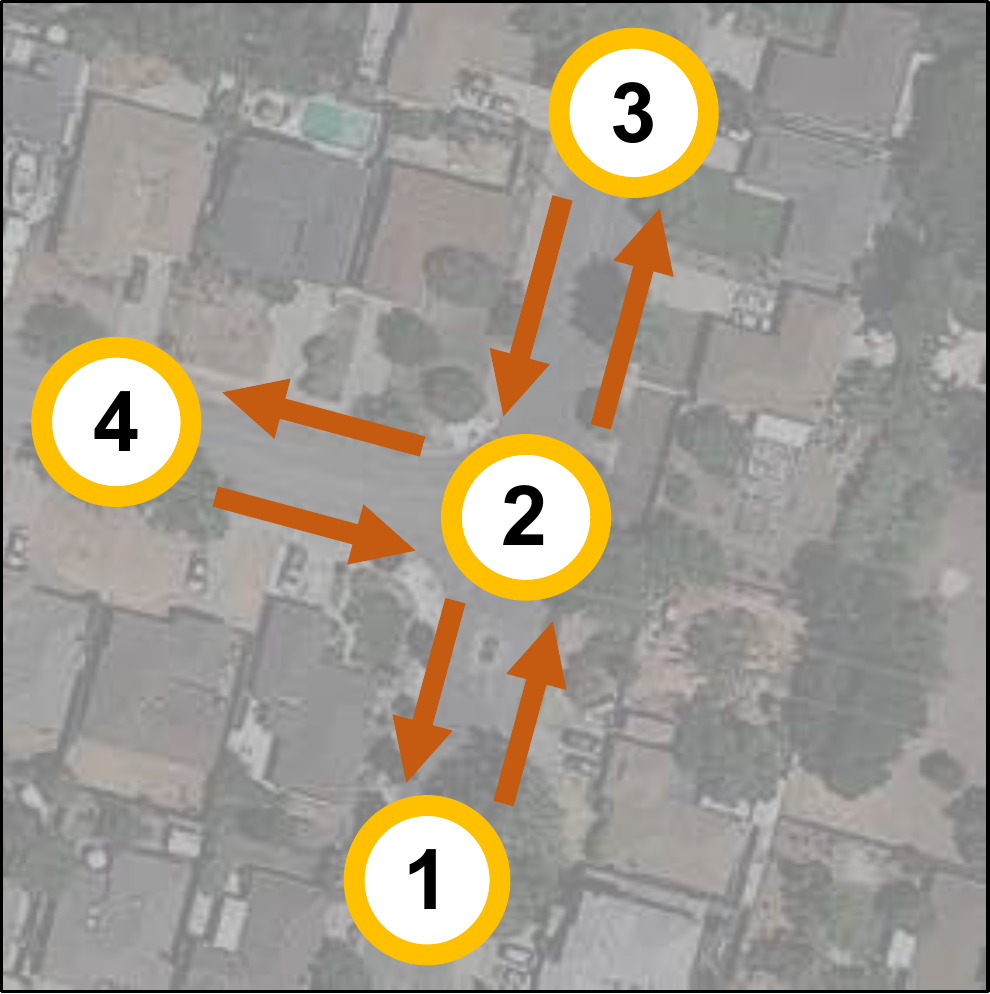}
        \caption{}
        \label{fig:path_polygon}
    \end{subfigure}
    \caption{Maze wall follower approach to sequentialization of road topology. (a) example aerial view of a T-junction, (b) wall follower sequence, (c) resulting ``polygon'' with sequence order 1$\to$2$\to$3$\to$2$\to$4$\to$2$\to$1.}
    \label{fig:path}
    \vspace{-1.5em}
\end{figure}

The inherent topology of objects such as roads or rivers is a general graph instead of a polygon, and the vertices of this graph are not necessarily connected in a sequential manner. In order to reformulate the topology of these objects as a polygon, we follow the principle of a maze solving algorithm, the wall follower, which is also known as the left-/right-hand rule (see Fig.~\ref{fig:path}): if a maze is simply connected, then by keeping one hand in contact with one wall of the maze, the algorithm is guaranteed to reach an exit.

We apply this principle to extract road sequences. As shown in Fig.~\ref{fig:path}, the road network can be regarded as a bidirected graph. Each road segment has two directed edges with opposite directions. We assume that for a given pair of directed edges, an edge's partner is always on its left when facing the direction of travel. Suppose we are standing at an arbitrary edge and we travel according to the following rules: (1) always walk facing the direction of the edge; (2) turn right when encountering an intersection; (3) turn around when encountering a dead end. Following this set of rules, we arrive back at the starting point after completing a full cycle (see Fig.~\ref{fig:path_true}). Finally, we connect all keypoints on the way (\textit{i.e.}, intersections and dead ends) in the order of traveling in order to obtain a ``polygon'' (see Fig.~\ref{fig:path_polygon}). In this way, the vertices that are originally not sequential in the road graph become ordered.

\begin{figure}[!ht]
	\centering
    \includegraphics[width=\columnwidth]{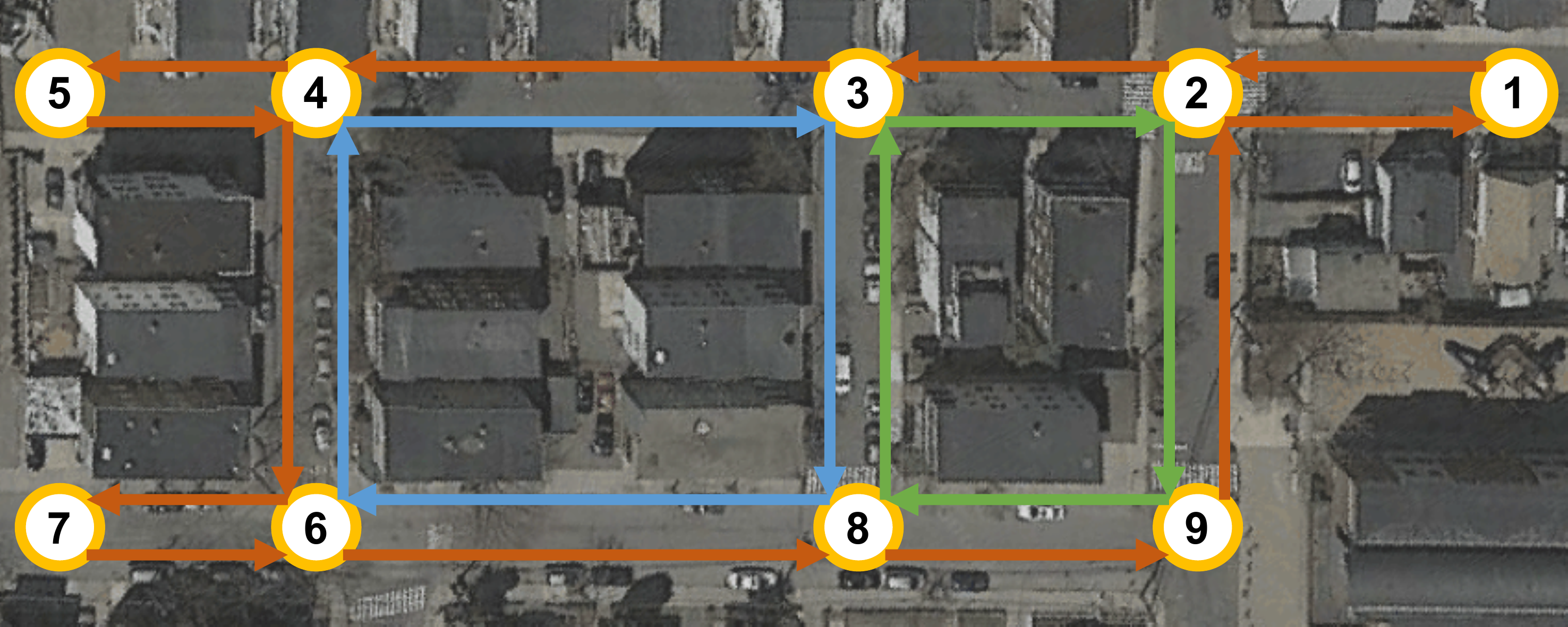}
    \caption{Road polygon extraction for a larger patch leading to one outer anticlockwise polygon (orange) and two inner clockwise polygons (blue and green).}
    \label{fig:maze_large}
    \vspace{-1.5em} 
\end{figure}

With a larger patch size or denser road networks, multiple polygons can exist as shown in Fig.~\ref{fig:maze_large}. However, we can only get a single polygon by following the rules described above. In order to get all the polygons in a graph, we need to traverse all the road segments twice (forward and backward). In practice, the sequence generation procedure goes as follows: we first traverse all edges in an arbitrary polygon, and for the directed edges that were not visited, we randomly select one and traverse it following the set of rules until all edges in the graph have been visited.

\subsection{Pipeline}
\label{sec:pipeline}

\begin{figure*}[!ht]
	\centering
    \includegraphics[width=\textwidth]{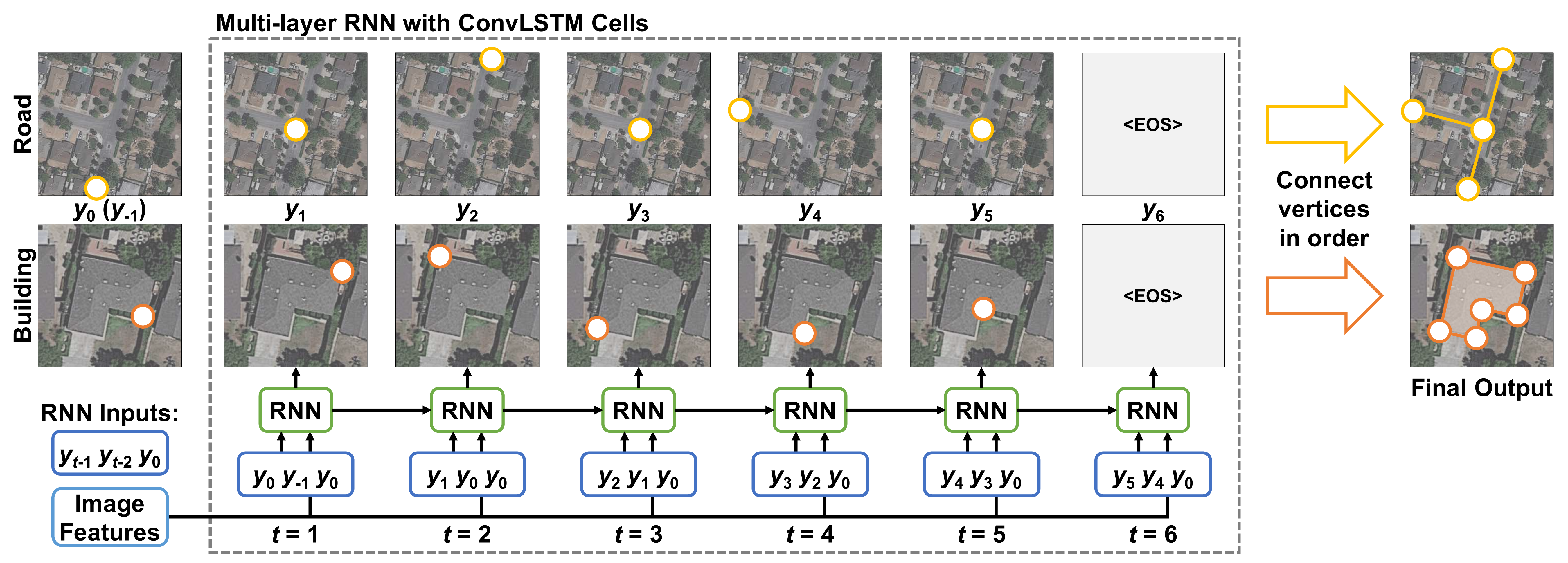}
    \caption{Keypoint sequence prediction produced by RNN for buildings and roads. At each time step $t$, the RNN takes the current vertex $y_t$ and previous vertex $y_{t-1}$ as input, as well as the first vertex $y_0$, and outputs a conditional probability distribution $P(y_{t+1} | y_t, y_{t-1}, y_0)$. When the polygon reaches its starting keypoint and becomes a closed shape, the end signal \lstinline{<eos>} is raised. Note that the RNN also takes features generated by the CNN (see Fig.~\ref{fig:cnn_building_road}) as input at each time step.}
    \label{fig:rnn}
    \vspace{-1.5em}
\end{figure*}

\paragraph{CNN Part}
For an input image, we first use a VGG-16 without tail layers as the CNN backbone to extract skip features~\cite{pinheiro2016learning} with $\frac{1}{8}$ the size of the input image (see Fig.~\ref{fig:cnn_building_road}). Meanwhile, the FPN also takes features from different layers of the backbone to construct a feature pyramid and predicts multiple bounding boxes containing the buildings.

For a single building, with the skip feature map and its bounding box, followed by RoIAlign~\cite{he2017mask}, the local features $F$ are obtained. We apply convolutional layers to the feature in order to generate a heat-map mask of building boundaries $B$ that delineate the object of interest. This is followed by additional convolutional layers outputting a mask of candidate keypoints, denoted by $V$. Both $B$ and $V$ have a size equal to $\frac{1}{8}$ the size of the input image. Among all candidate keypoints, we select those $w$ points with the highest score in $V$ as starting point $y_0$ (same as $y_{-1}$, see Fig.~\ref{fig:rnn}).

As illustrated in Fig.~\ref{fig:cnn_building_road}, the main procedure of road network extraction is identical to the case of buildings. We only adapt RoI definition and vertex selection to the road case. While building RoIs are sampled within an image patch, a road RoI corresponds to the entire image patch. Naturally, the generated heatmap $B$ refers to the roads' centerlines instead of building boundaries. Vertex selection is adapted to the road topology by selecting start point candidates at image edges and choosing the one with the highest score as starting point $y_0$ (same as $y_{-1}$) to predict the unique outer polygon. Note that each segment of the outer polygon should be passed twice unless the segment is shared with an inner polygon. Thus, after the outer polygon is predicted, we choose two vertices of a segment that is passed only once as $y_{-1}$ and $y_0$ (in reverse direction) to further predict a potential inner polygon.

\vspace{-1em}
\paragraph{RNN Part}
As illustrated in Fig.~\ref{fig:rnn}, the RNN outputs $y_t$'s potential location $P(y_{t+1} | y_t, y_{t-1}, y_0)$ at each step $t$. We input both, $y_t$ and $y_{t-1}$ to compute the conditional probability distribution of $y_{t+1}$ because it allows defining a unique direction. If given two neighboring vertices with an order in a polygon, the next vertex in this polygon is uniquely determined. Note that the distribution also involves the end signal \lstinline{<eos>} (end of sequence), which indicates that the polygon reaches a closed shape and the prediction procedure should come to the end. The final, end vertex in a polygon thus corresponds to the very first, starting vertex $y_0$, which therefore has to be included at each step.

In practice, we ultimately concatenate $F$, $B$, $V$, $y_0$ (also $y_{-1}$ for polygon prediction in the case of roads) and feed the resulting tensor to a multi-layer RNN with ConvLSTM~{\cite{convlstm}} cells in order to sequentially predict the vertices that will delineate the object of interest, until it predicts the \lstinline{<eos>} symbol. For buildings, we simply connect all sequentially predicted vertices to obtain the final building polygon. In the case of roads, the predicted polygon(s) themselves are not needed directly but rather used as a set of edges between vertices. We thus use all these individual line segments that make up the polygon(s) for further processing. Specifically, each of the predicted segments $\mathbf{e}$ is associated with a score $s_{\mathbf{e}}$ calculated as $s_{\mathbf{e}} = \int_{0}^{1} B(\mathbf{e}(u)) \text{d}u \in [0, 1]$, where $\mathbf{e}(u) = ue_{1} + (1-u)e_{2}$, $B$ is the heatmap of centerlines, $e_{1}$ and $e_{2}$ are the two extremities of $\mathbf{e}$. We remove segments with low scores and connect the remaining segments to form the entire graph.

\subsection{Implementation Details}
\label{subsec:impl}

We set the model parameters using size 28$\times$28 for $F$, $B$, $V$ and $y_t$, and set the number of layers of the RNN to 3 (buildings) and 4 (roads). The maximum length of a sequence when training is set to be 30 for both cases. The total loss of the building case is a combined loss from the FPN, CNN and RNN parts. The FPN loss consists of a cross-entropy loss for anchor classification and a smooth L1 loss for anchor regression. The CNN loss refers to the log loss for the mask of boundary and vertices, and the RNN loss is the cross-entropy loss for the multi-class classification at each time step. In the road case, the FPN loss is excluded.

For training, we use the Adam~\cite{kingma14adam} optimizer with batch size 4 and an initial learning rate of 0.0001, as well as default $\beta_\text{1}$ and $\beta_\text{2}$. We trained our model on 4 GPUs for a day for buildings and 12 hours for roads. During training, we force the order in which we visit the edges of the building polygons to be anticlockwise, while for the road polygons we follow the set of rules described in Sec.~\ref{sec:roads}.

In the inference phase, we use beam search with a width $w$ (which is 5 in our experiments). For building, we select top $w$ vertices with highest probability in $V$ as the starting vertices, then followed by a general beam search procedure. Among the $w$ polygon candidates, we choose the one with the highest probability as the output. Similarly, for road, we select vertices at the edge of the image and then choose top $w$ with the highest score as the starting point and follows the general beam search algorithm. After the outer polygon is predicted, we can further predict potential inner polygon(s) as mentioned in Sec.~\ref{sec:pipeline}. Finally, we use a threshold of 0.7 (which was found to yield good results) in our experiments to exclude unmatched edges.

In addition, for the topological map extraction from a relatively large-scale overhead image of a city, we first divide the whole image into several patches with 50\% coverage. In the training phase of the building footprints, incomplete footprints at the edge of the image are still be used, however, they are excluded in the inference scheme. In the case of roads, in order to get a complete city road network, some post-processing is performed, such as splicing road networks in adjacent patches, removing small loops of the graph and duplicated vertices and edges.

As for the efficiency, the average inference time on a single GPU is 0.38s for buildings and 0.29s for roads per image patch (300$\times$300 pixels).
\section{Experiments}
\label{sec:exp}

\begin{table*}[!ht]
    \centering
	\caption[Building footprint extraction results on the crowdAI dataset~\cite{crowdAIdataset}]{Buildings extraction results on the crowdAI dataset~\cite{crowdAIdataset}}
	\label{tab:eval_crowdai}
	\begin{tabular}{r|r r r r r r|r r r r r r}
	\hline
	\textbf{Method} & AP & AP$_{\text{50}}$ & AP$_{\text{75}}$ & AP$_{\text{S}}$ & AP$_{\text{M}}$ & AP$_{\text{L}}$ & AR & AR$_{\text{50}}$ & AR$_{\text{75}}$ & AR$_{\text{S}}$ & AR$_{\text{M}}$ & AR$_{\text{L}}$ \\ \hline
	Mask R-CNN\cite{he2017mask, crowdAIMappingChallengeBaseline2018} & 41.9 & 67.5 & 48.8 & 12.4 & 58.1 & 51.9 & 47.6 & 70.8 & 55.5 & 18.1 & 65.2 & 63.3 \\
	PANet\cite{liu2018path} & 50.7 & 73.9 & 62.6 & 19.8 & \textbf{68.5} & \textbf{65.8} & 54.4 & 74.5 & 65.2 & 21.8 & 73.5 & 75.0 \\ \hline
	PolyMapper & \textbf{55.7} & \textbf{86.0} & \textbf{65.1} & \textbf{30.7} & \textbf{68.5} & 58.4 & \textbf{62.1} & \textbf{88.6} & \textbf{71.4} & \textbf{39.4} & \textbf{75.6} & \textbf{75.4} \\ \hline
    \end{tabular}
\end{table*}

\begin{figure*}[!ht]
	\centering
	\begin{subfigure}{0.3302\textwidth}
	    \centering
        \includegraphics[width=\textwidth]{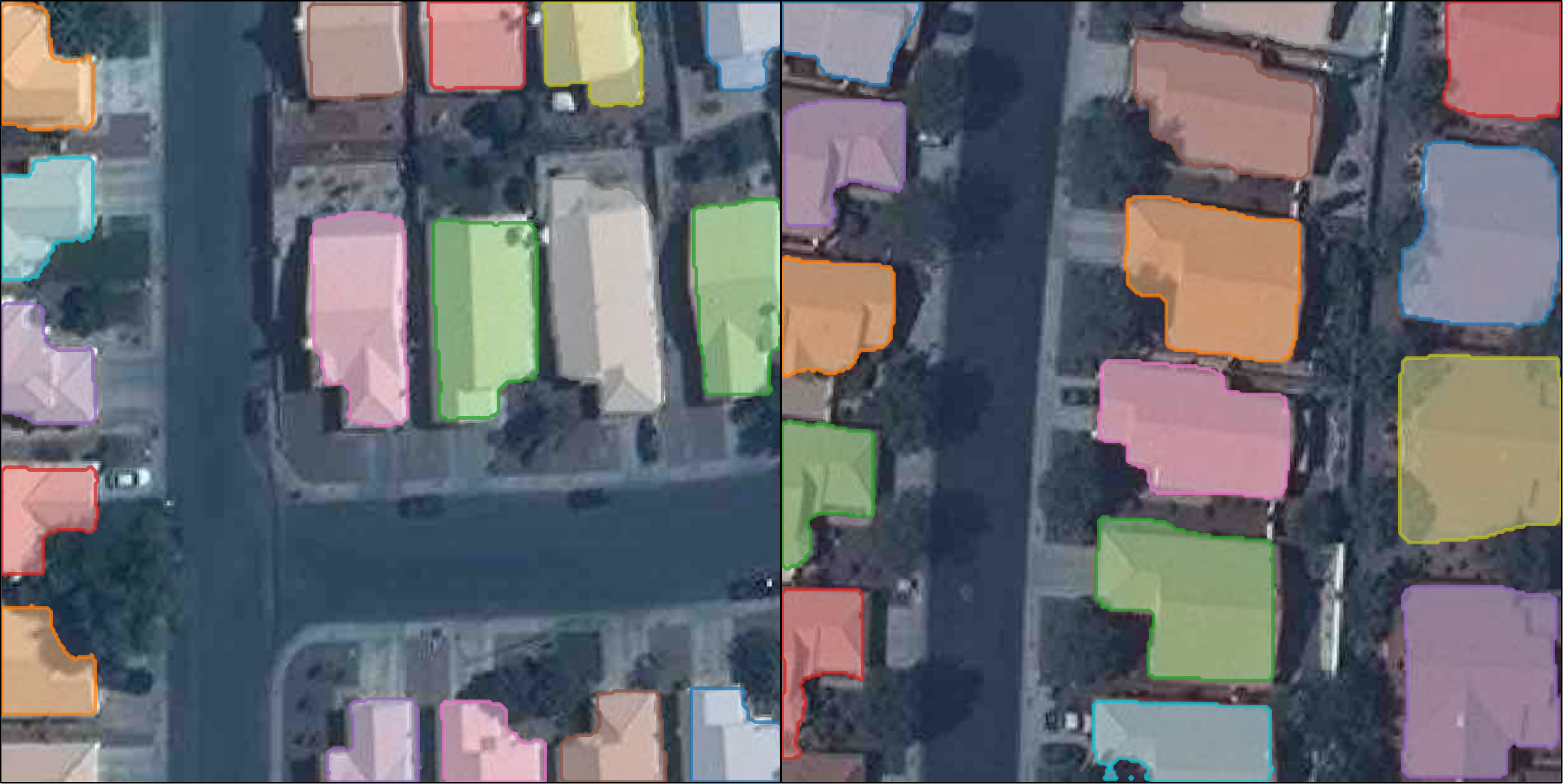}
        \caption{Mask R-CNN~\cite{he2017mask, crowdAIdataset}}
        \label{fig:building_maskrcnn}
    \end{subfigure}
    \begin{subfigure}{0.3302\textwidth}
        \centering
        \includegraphics[width=\textwidth]{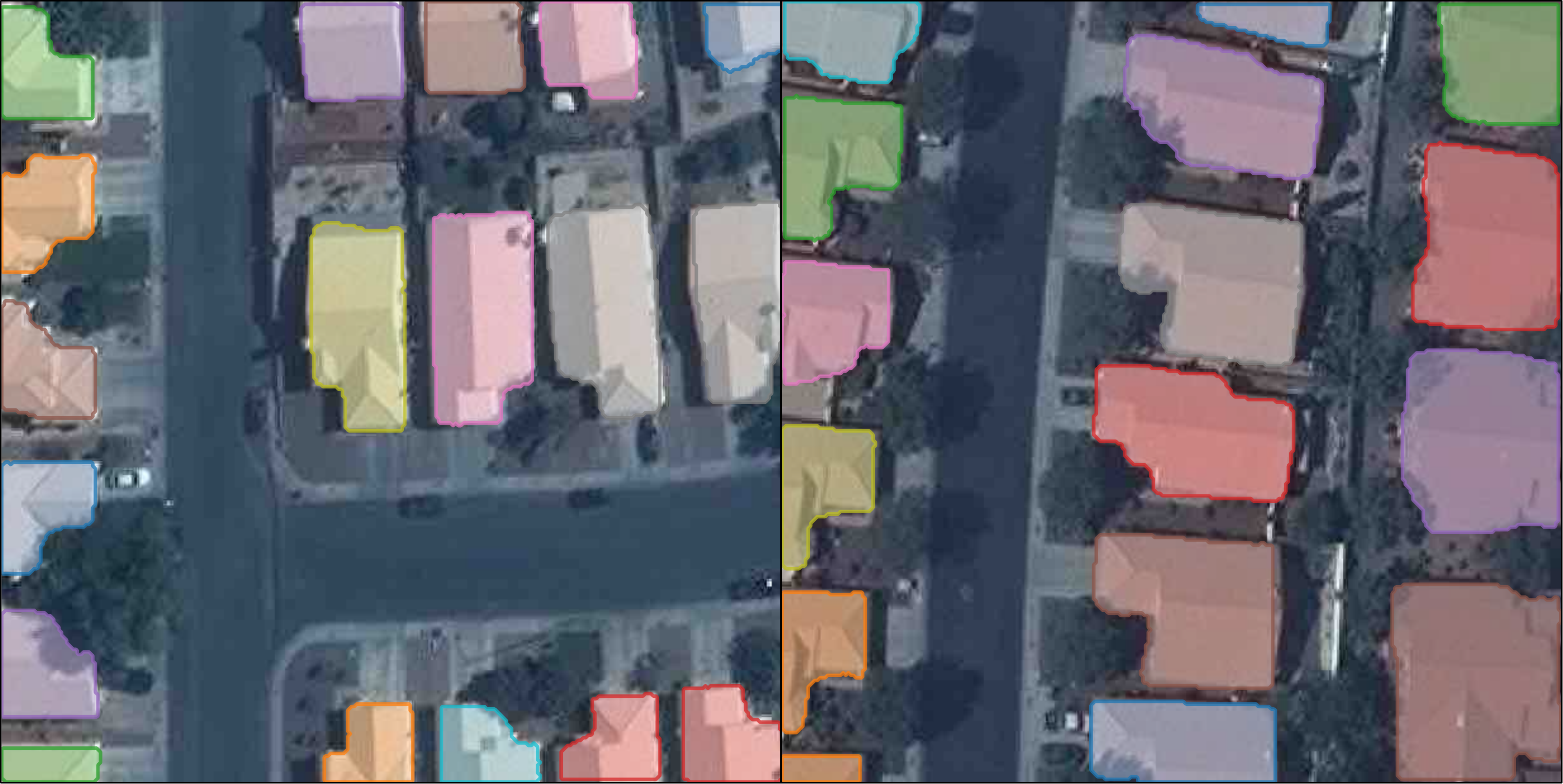}
        \caption{PANet~\cite{liu2018path}}
        \label{fig:building_panet}
    \end{subfigure}
    \begin{subfigure}{0.3302\textwidth}
        \centering
        \includegraphics[width=\textwidth]{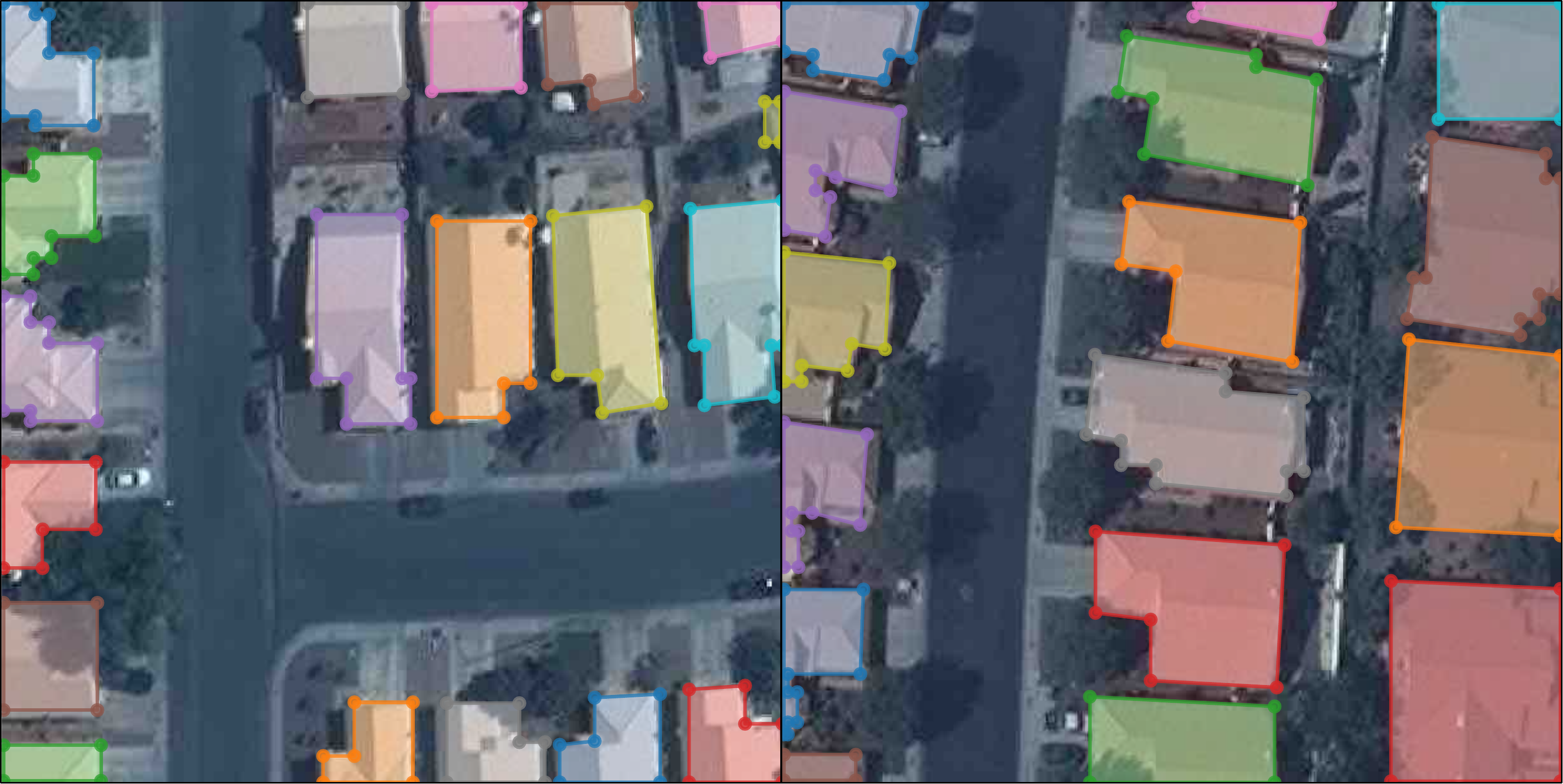}
        \caption{PolyMapper}
        \label{fig:building_ours}
    \end{subfigure}
    \caption{Building footprint extraction results on 2 example patches of the crowdAI dataset~\cite{crowdAIdataset} achieved with (a) Mask R-CNN~\cite{he2017mask, crowdAIdataset}, (b) PANet~\cite{liu2018path}, and (c) PolyMapper. Note that results in (a) and (b) are images labeled per pixel whereas PolyMapper shows polygons, as well as vertices connected with line segments.}
    \label{fig:building_vis}
    \vspace{-1.5em}
\end{figure*}

\begin{figure}[!ht]
    \centering
    \begin{subfigure}{0.3269\columnwidth}
	    \centering
        \includegraphics[width=\textwidth]{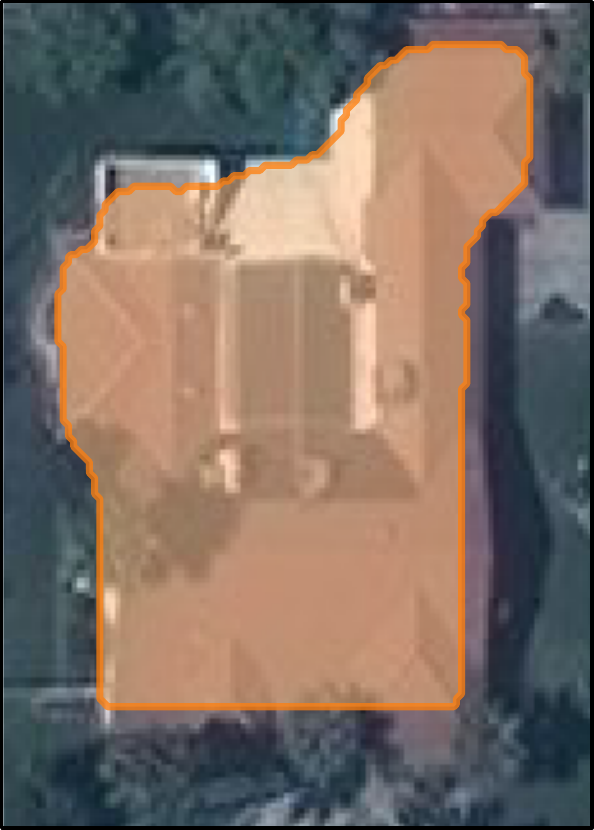}
        \caption{Mask R-CNN}
        \label{fig:cut_maskrcnn}
    \end{subfigure}
    \begin{subfigure}{0.3269\columnwidth}
	    \centering
        \includegraphics[width=\textwidth]{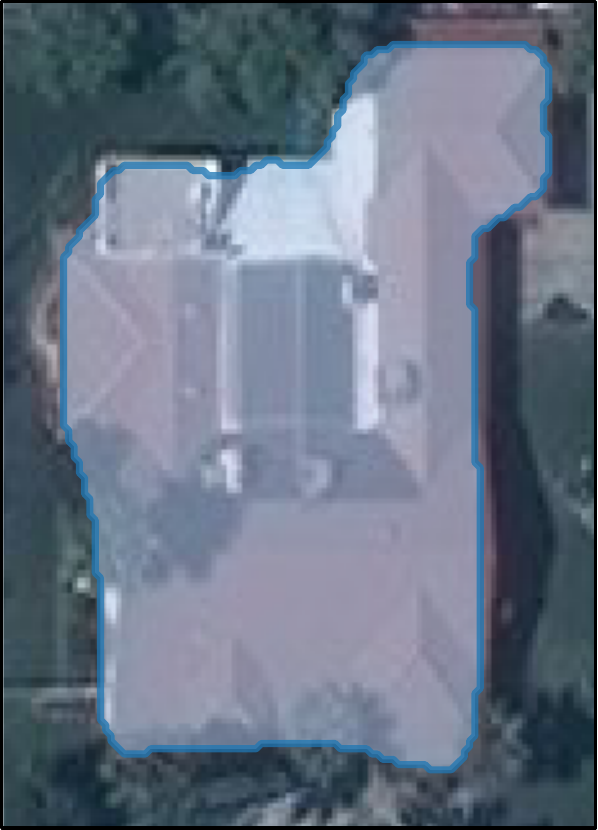}
        \caption{PANet}
        \label{fig:cut_panet}
    \end{subfigure}
    \begin{subfigure}{0.3269\columnwidth}
	    \centering
        \includegraphics[width=\textwidth]{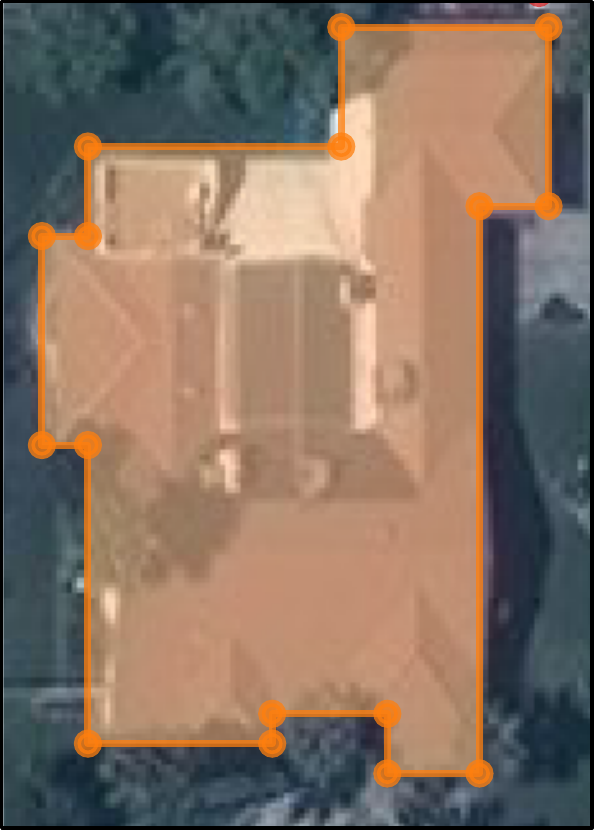}
        \caption{PolyMapper}
        \label{fig:cut_ours}
    \end{subfigure}
    \caption{Comparison of pixel-wise semantic segmentation results of Mask R-CNN and PANet with our direct polygon prediction PolyMapper for an example building.}
	\label{fig:building_cut}
	\vspace{-1.5em}
\end{figure}

We are not aware of any publicly available dataset\footnote{Note that the only dataset that contains both, building footprints and road centerlines is SpaceNet~\cite{vanetten2018}, which runs on the Amazon Cloud and uses images of lower resolution than ours. In addition, we are not aware of any scientific publication of a state-of-the-art approach that uses it.} that contains labeled building footprints and road networks together with aerial images at large scale and thus create our own dataset (see Sec.~\ref{sec:res_ours}). In order to compare our results to the state-of-the-art, we resort to evaluating building footprint extraction and road network delineation separately on popular task-specific datasets, crowdAI \cite{crowdAIdataset} and RoadTracer \cite{bastani2018roadtracer} (see Sec.~\ref{sec:comp_star}).

\subsection{Evaluation Measures}
\label{sec:eval}

For building extraction, we report the standard MS COCO measures including average precision (AP, averaged over IoU thresholds), AP$_{\text{50}}$, AP$_{\text{75}}$ and AP$_{\text{S}}$, AP$_{\text{M}}$, AP$_{\text{L}}$ (AP at different scales). To measure the proportion of buildings detected by our approach with respect to the ground truth, we additionally evaluate average recall (AR), which is not commonly used in previous works such as~\cite{he2017mask, liu2018path}. Both AP and AR are evaluated using mask IoU. However, we would like to emphasize that in contrast to pixel-wise output masks produced by common methods for building footprint extraction, our outputs are polygon representations of building footprints.

Evaluating the quality of road networks in terms of its topology is a non-trivial problem. \cite{wegner13road} propose a connectivity measure SP, which centers on evaluating shortest path distances between randomly chosen point pairs in the road graph. SP generates a large number of pairs of vertices, computes the shortest path between each two vertices in both ground truth and predicted maps, and outputs the fraction of pairs where the predicted length is equal (up to a buffer of 10\%) to the ground truth, shorter (erroneous shortcut) or longer (undetected piece of road).

In addition to SP, we propose a new topology evaluation measure that compares shortest paths through graphs \cite{wegner13road} using a measure based on average precision (AP) and average recall (AR). This allows an evaluation similar to building footprints and compares ground truth and predicted road graphs in a meaningful way. Similar to the definition in \cite{deeproadmapper}, we define the similarity score for the length of two shortest paths, $d^*$ and $d$, in ground truth and predicted road graphs as a ratio of minimum and maximum values,
\begin{equation}
    \text{IoU}(d^*, d) = \text{IoU}(d, d^*) = \frac{\text{min}(d^*,d)}{\text{max}(d^*,d)} \in [0, 1].
\end{equation}
Then, with a given IoU threshold $t$, we can define the weighted precision and recall as follows,
\begin{equation}
\text{AP}^{\text{IoU}=t} = \frac{\sum_{i} d_i \mathbbm{1}[\text{IoU}(d_i, d^*_{j_i}) \geq t]}{\sum_{i} d_i},
\end{equation}
\begin{equation}
\text{AR}^{\text{IoU}=t} = \frac{\sum_{j} d^*_j \mathbbm{1}[\text{IoU}(d^*_j, d_{i_j}) \geq t]}{\sum_{j} d^*_j},
\end{equation}
where $\mathbbm{1}[\cdot]$ is the indicator function, $d_i$ and $d^*_{j_i}$ refer to the $i$-th shortest path in the inferred map and its corresponding shortest path with index $j_i$ in the ground truth graph, similar for $d^*_j$ and $d_{i_j}$. Note that the shortest path computation is expensive and it is unfeasible to compute all possible paths exhaustively. We thus randomly sample 100 start vertices and sample 1,000 end vertices for each of them, which yields 100,000 shortest paths in total.

\begin{table*}[!ht]
    \centering
	\caption[Road network extraction results on the RoadTracer dataset~\cite{bastani2018roadtracer}]{Road network extraction results on the RoadTracer dataset~\cite{bastani2018roadtracer}}
	\label{tab:eval_roadtracer}
	\begin{tabular}{r|r r|r r r|r r r}
	\hline
	\textbf{Method} & SP$_{\pm\text{5\%}}$ & SP$_{\pm\text{10\%}}$ & AP$_{\text{85}}$ & AP$_{\text{90}}$ & AP$_{\text{95}}$ & AR$_{\text{85}}$ & AR$_{\text{90}}$ & AR$_{\text{95}}$ \\ \hline
	DeepRoadMapper~\cite{deeproadmapper} & 11.9 & 15.6 & 35.9 & 28.4 & 19.1 & 58.2 & 45.7 & 27.8 \\
	RoadTracer~\cite{bastani2018roadtracer} & \textbf{47.2} & \textbf{61.8} & 64.9 & 56.6 & \textbf{42.4} & \textbf{85.3} & \textbf{76.5} & \textbf{56.8} \\ \hline
	PolyMapper & 45.7 & 61.1 & \textbf{65.5} & \textbf{57.2} & 40.7 & 84.2 & 74.8 & 53.7 \\ \hline
    \end{tabular}
\end{table*}

\begin{figure*}[!ht]
	\centering
	\begin{subfigure}{0.3302\textwidth}
	    \centering
        \includegraphics[width=\textwidth]{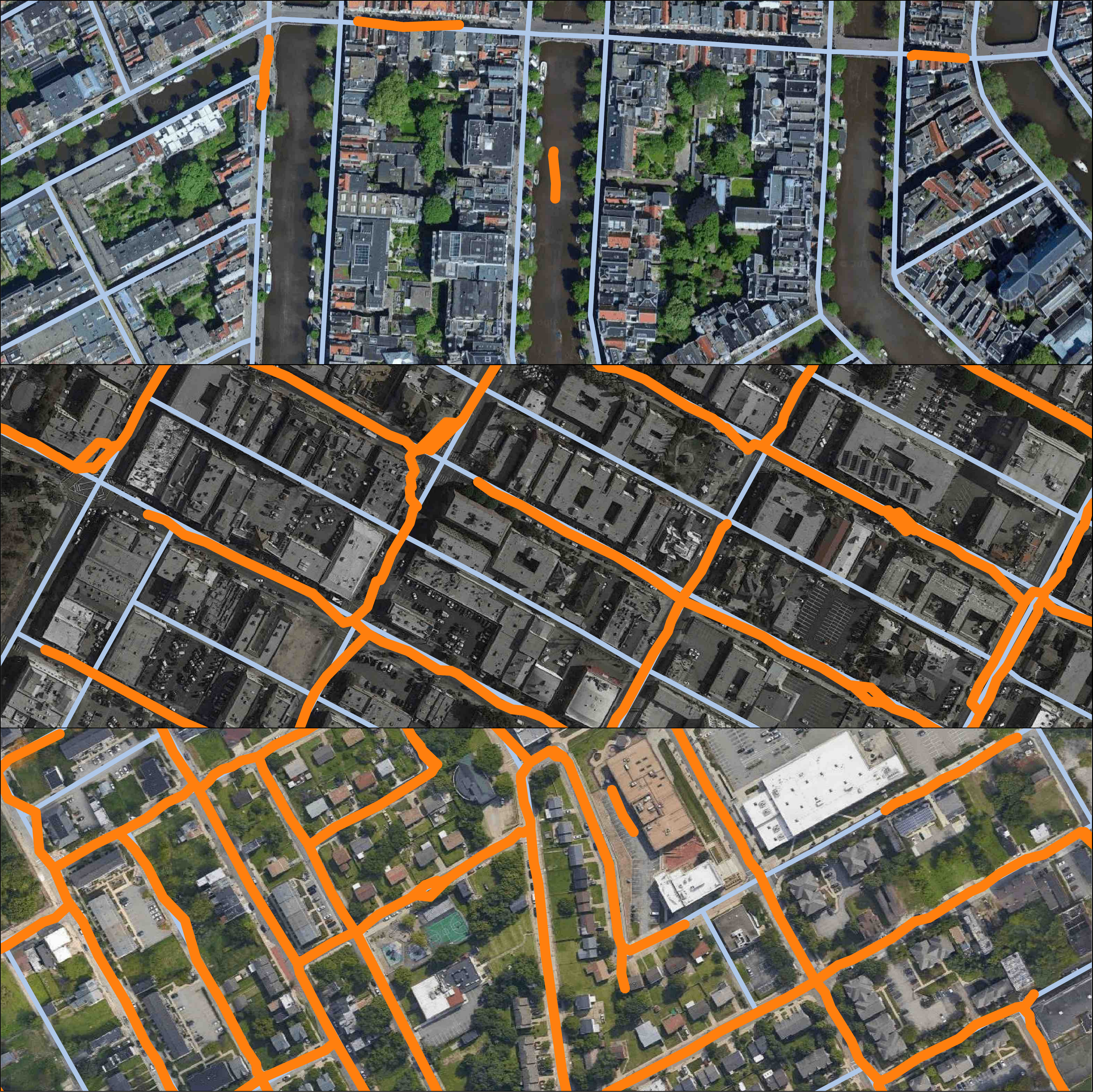}
        \caption{DeepRoadMapper\cite{deeproadmapper}}
        \label{fig:road_drm}
    \end{subfigure}
    \begin{subfigure}{0.3302\textwidth}
        \centering
        \includegraphics[width=\textwidth]{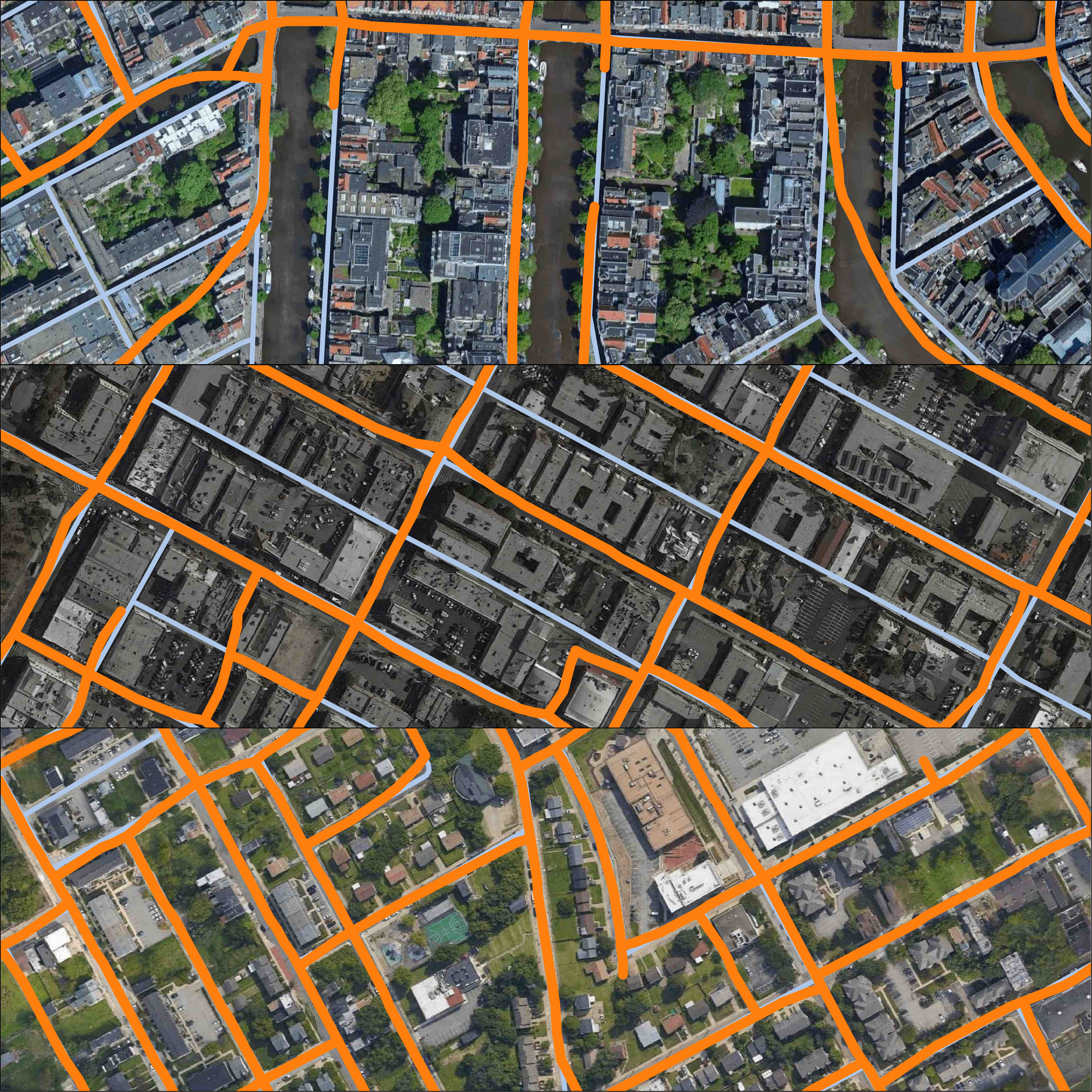}
        \caption{RoadTracer\cite{bastani2018roadtracer}}
        \label{fig:road_rt}
    \end{subfigure}
    \begin{subfigure}{0.3302\textwidth}
        \centering
        \includegraphics[width=\textwidth]{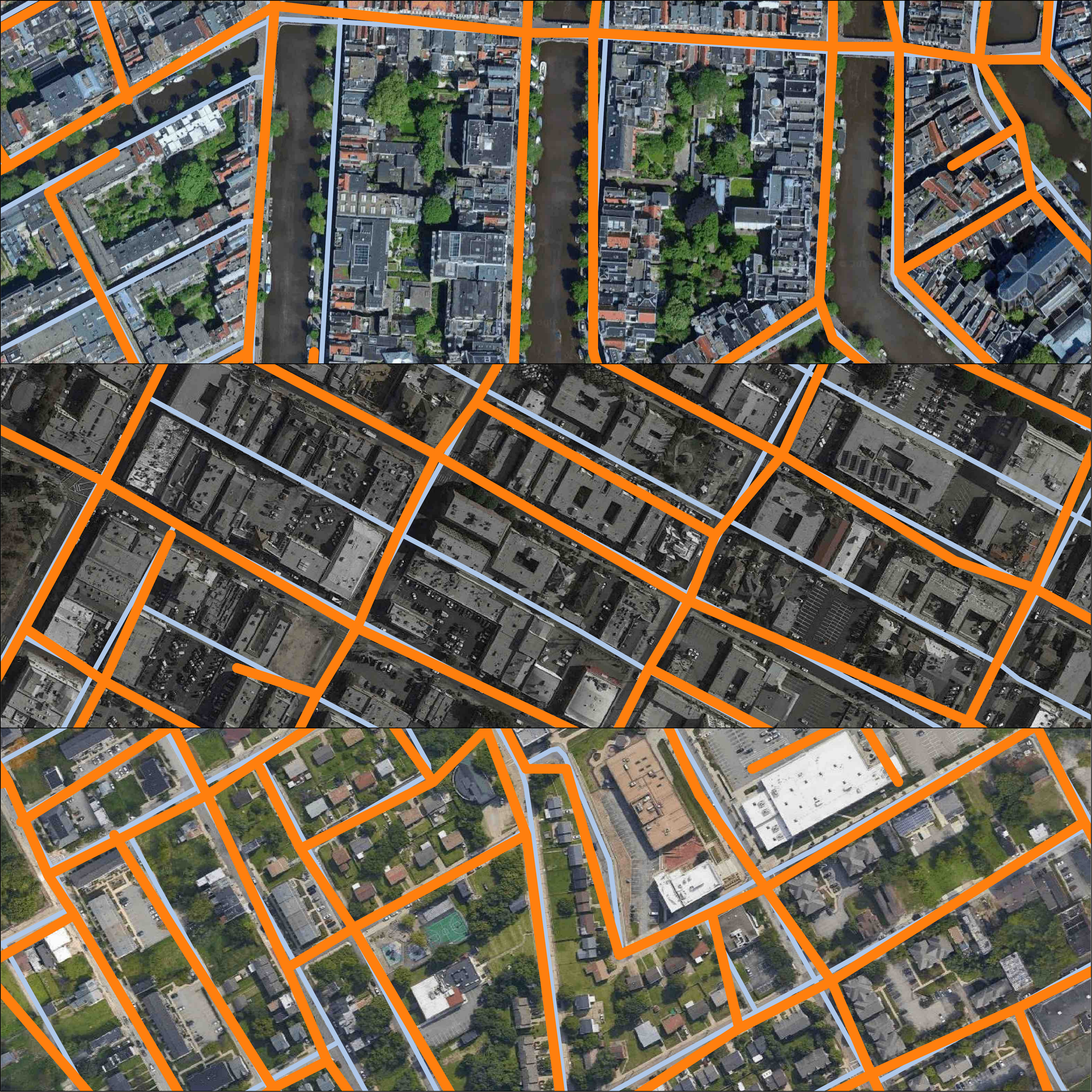}
        \caption{PolyMapper}
        \label{fig:road_ours}
    \end{subfigure}
    \caption{Comparison of predicted road network (orange) to ground truth (blue) for subscenes of Amsterdam (top), Los Angeles (middle) and Pittsburgh (bottom) of the RoadTracer dataset~\cite{bastani2018roadtracer}.}
    \label{fig:road_vis}
    \vspace{-1.5em}
\end{figure*}

\begin{figure}[!ht]
    \centering
    \begin{subfigure}{0.3269\columnwidth}
	    \centering
        \includegraphics[width=\textwidth]{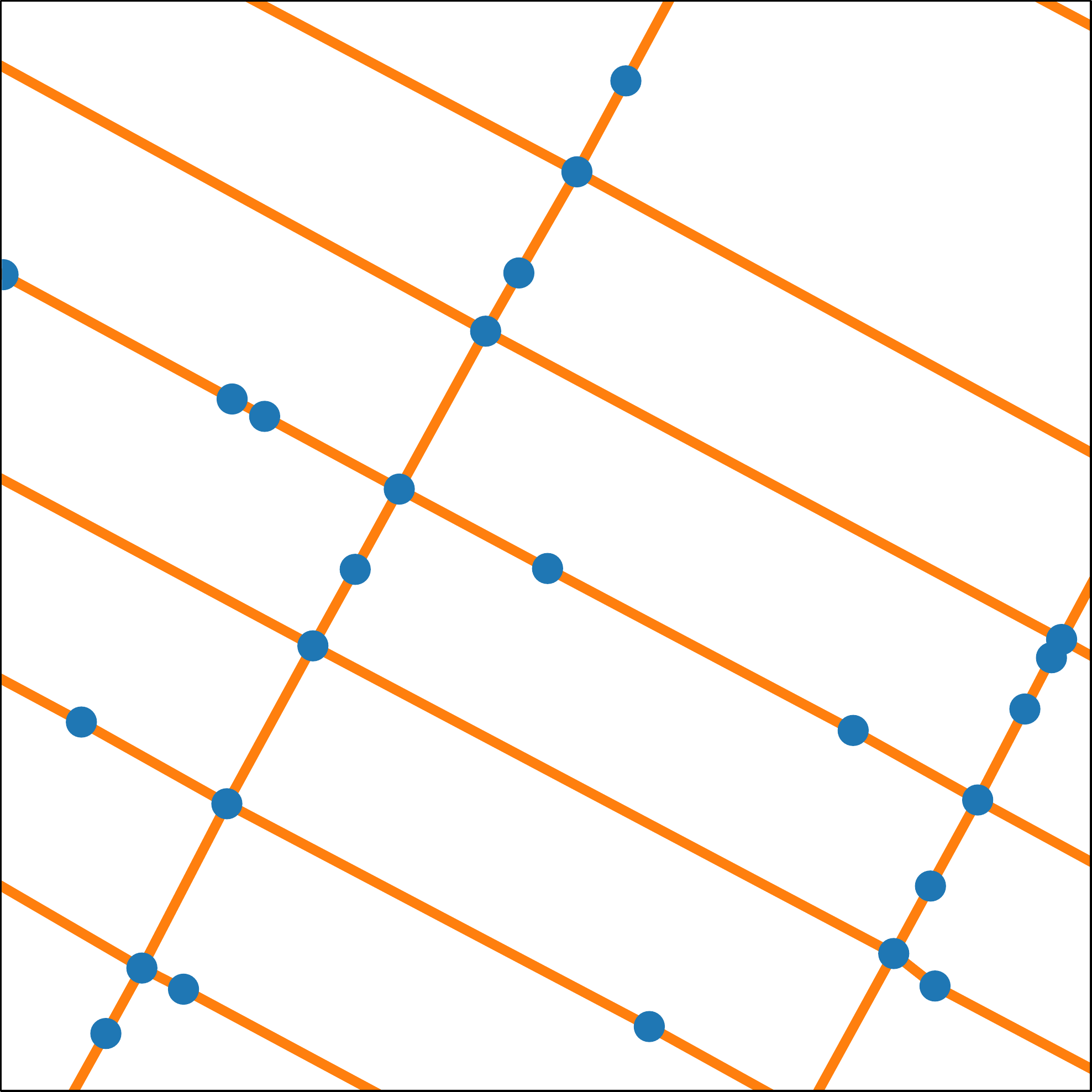}
        \caption{Ground Truth}
        \label{fig:v_gt}
    \end{subfigure}
    \begin{subfigure}{0.3269\columnwidth}
	    \centering
        \includegraphics[width=\textwidth]{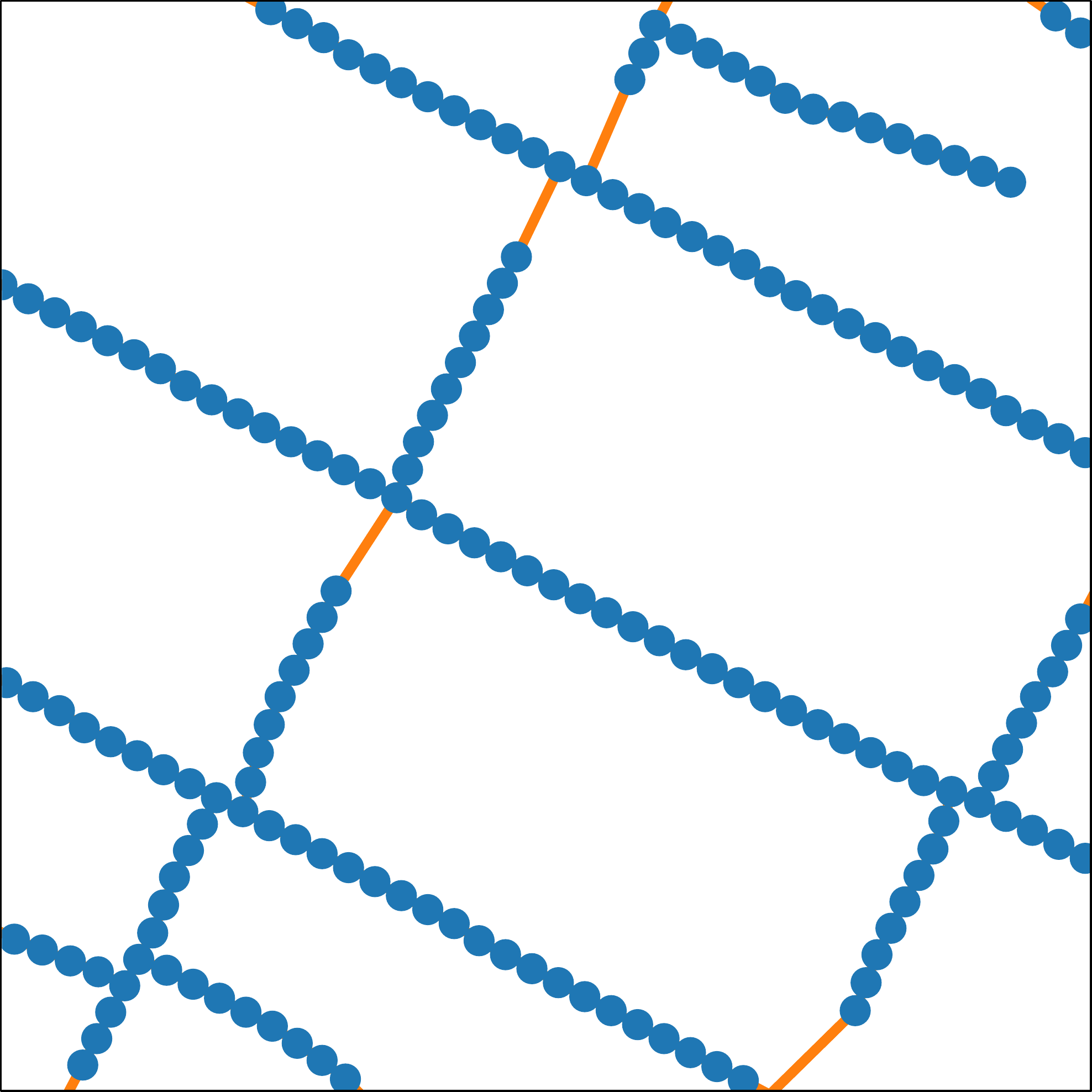}
        \caption{RoadTracer~\cite{bastani2018roadtracer}}
        \label{fig:v_rt}
    \end{subfigure}
    \begin{subfigure}{0.3269\columnwidth}
	    \centering
        \includegraphics[width=\textwidth]{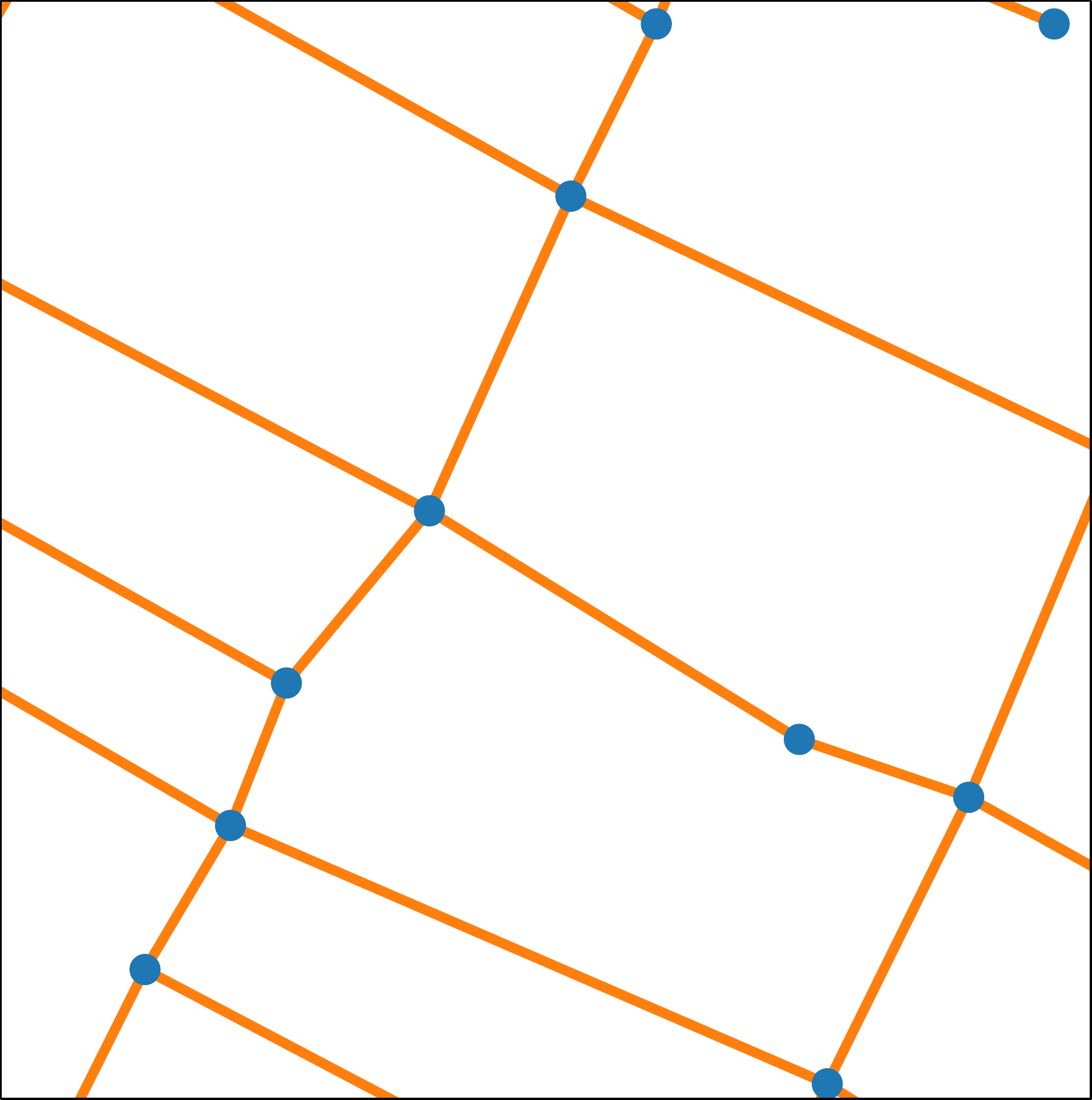}
        \caption{PolyMapper}
        \label{fig:v_ours}
    \end{subfigure}
    \caption{Visual comparison of graph structures. Vertices are blue and edges are orange.}
	\label{fig:v}
	\vspace{-1.5em}
\end{figure}

\subsection{Comparison to State-of-the-art}
\label{sec:comp_star}

\paragraph{Buildings}
We use the crowdAI dataset \cite{crowdAIdataset} to validate the building footprint extraction results and to compare to the state-of-the-arts. This large-scale dataset is split as follows. The training set consists of $\sim$280,000 images with $\sim$2,400,000 annotated building footprints. The test set contains $\sim$60,000 images with $\sim$515,000 buildings. Each individual building is annotated in a polygon format as a sequence of vertices according to MS COCO~\cite{mscoco} standards.

We compare the performance of our model on the crowdAI dataset~\cite{crowdAIdataset} to state-of-the-art methods Mask R-CNN~\cite{he2017mask, crowdAIMappingChallengeBaseline2018} and PANet \cite{liu2018path}. Results in Tab.~\ref{tab:eval_crowdai} show that PolyMapper outperforms Mask R-CNN and PANet in all AP and AR metrics except AP$_{\text{L}}$, which refers to large buildings. We hypothesize that the inferior performance observed for large buildings is due to their large feature maps, which leads to more inaccurate location information since a vertex location may be blurred when re-sizing to a fixed size.
Fig.~\ref{fig:building_vis} and Fig.~\ref{fig:building_cut} provides a qualitative comparison of the predictions of the state-of-the-art methods and PolyMapper, where polygons appear to be a more compact representation for buildings. We also see that PolyMapper learns to produce right angles on its own. As future work, we would like to explore whether imposing more geometrical constraints could further improve the results.

\vspace{-1em}
\paragraph{Roads}
To evaluate the road network extraction we use the dataset of~\cite{bastani2018roadtracer} tailored for the RoadTracer method. We used their code to download the entire dataset and we trained our model using the same train and test split. Note that we train and test on images from 25 and 15 cities respectively. Our results thus indicate to a certain extent how well an approach generalizes to new scenes.

We compare the results of our method to the state-of-the-art methods DeepRoadMapper~\cite{deeproadmapper} and RoadTracer~\cite{bastani2018roadtracer}. We directly take the predicted graphs for both models from \cite{bastani2018roadtracer} (who re-implemented \cite{deeproadmapper}) and compute evaluation measures SP, AP and AR as shown in Tab.~\ref{tab:eval_roadtracer}. A visual comparison of the results overlaid on top of the original images is shown in Fig.~\ref{fig:road_vis} whereas a comparison of the graph structures is shown in Fig.~\ref{fig:v}. PolyMapper outperforms DeepRoadMapper\cite{deeproadmapper} in all measures and performs on par with RoadTracer~\cite{bastani2018roadtracer}.

We visually compare the PolyMapper graph structure to ground truth and RoadTracer~\cite{bastani2018roadtracer} in Fig.~\ref{fig:v}. The road graph representation of PolyMapper is close to the ground truth whereas RoadTracer predicts many more vertices. We compare the overall graph complexity in terms of the total number of vertices and edges in Tab.~\ref{tab:comp_num} for 15 cities of the RoadTracer test set. PolyMapper has a much lower graph complexity with $\sim$87\% less vertices and edges than RoadTracer~\cite{bastani2018roadtracer} and $\sim$70\% less than DeepRoadMapper~\cite{deeproadmapper}.

\vspace{-0.5em}
\begin{table}[!ht]
    \centering
    \caption{Comparison of graph complexity}
    \begin{tabular*}{0.84\columnwidth}{r|r r}
    \hline
    \textbf{Method} & \#Vertices & \#Edges \\ \hline
    DeepRoadMapper~\cite{deeproadmapper} & 126,029 & 118,978 \\
    RoadTracer~\cite{bastani2018roadtracer} & 271,244 & 281,518 \\ \hline
    PolyMapper & \textbf{31,749} & \textbf{35,998} \\ \hline
    \end{tabular*}
    \label{tab:comp_num}
\end{table}
\vspace{-1em}

\begin{figure}[!ht]
    \centering
    \begin{subfigure}{\columnwidth}
        \centering
        \includegraphics[width=\columnwidth]{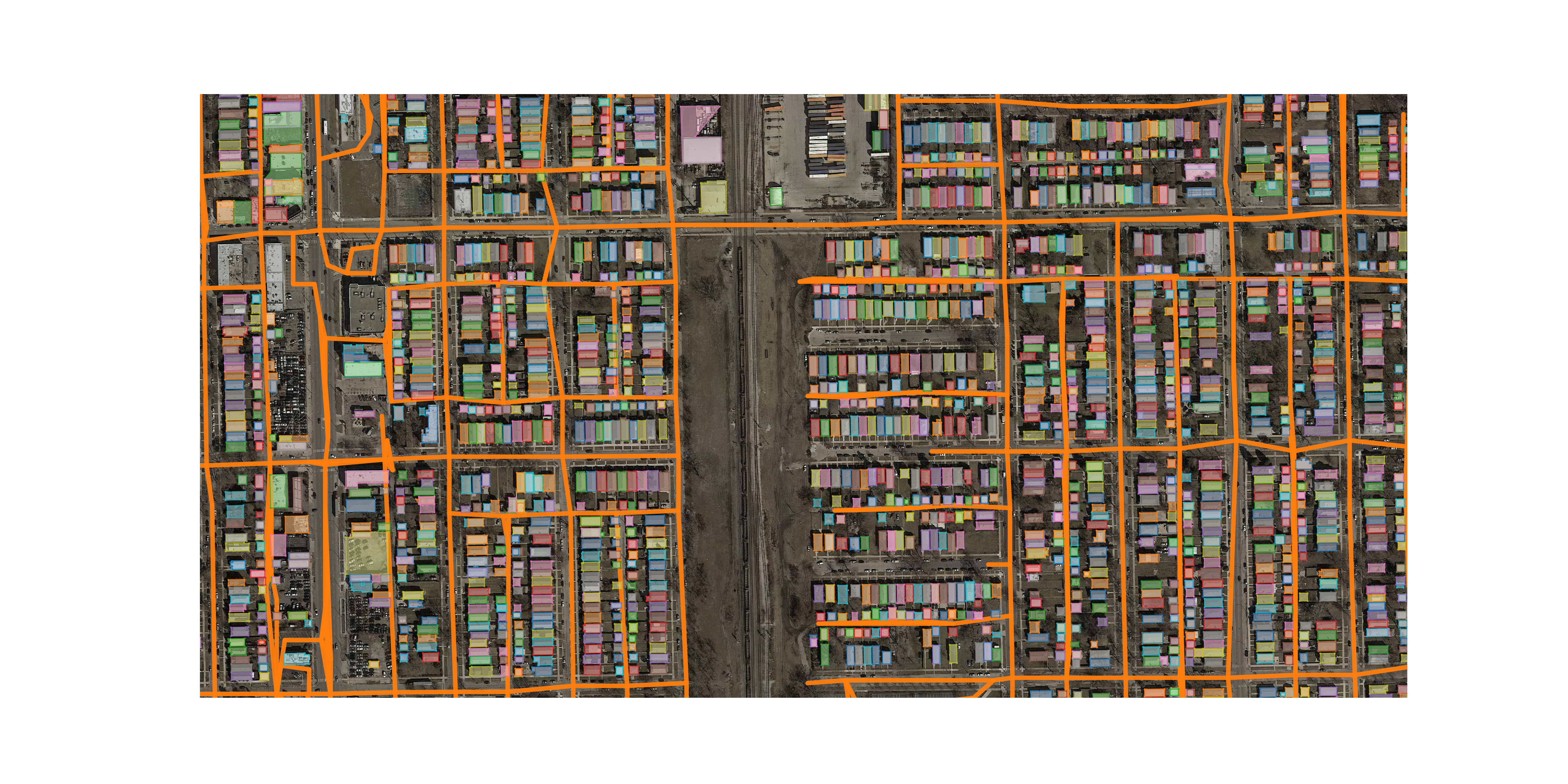}
        \caption{Chicago}
        \label{fig:demo_chicago}
    \end{subfigure}\\
    \begin{subfigure}{\columnwidth}
        \centering
        \includegraphics[width=\columnwidth]{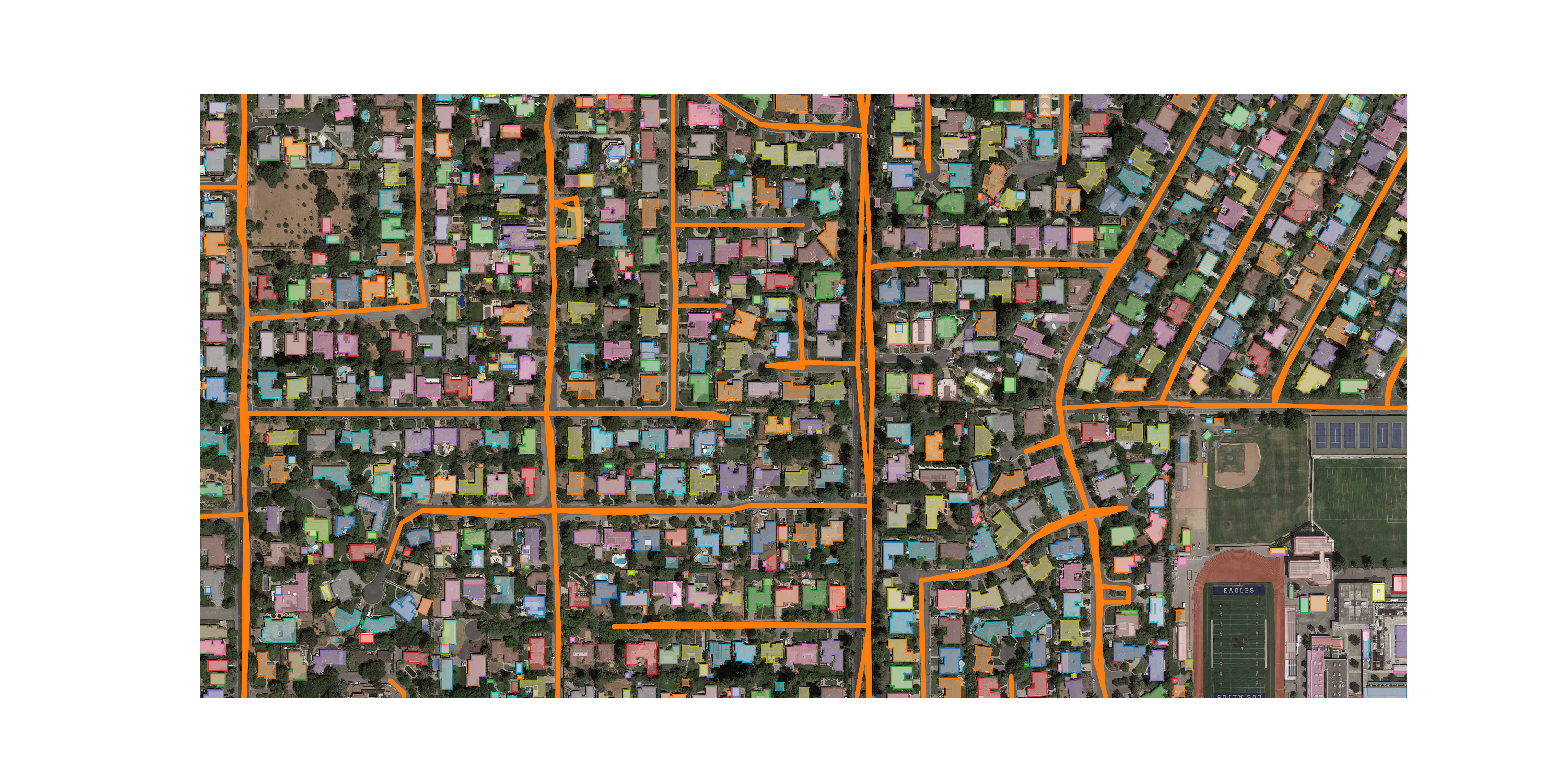}
        \caption{Sunnyvale}
        \label{fig:demo_sunnyvale}
    \end{subfigure}
    \caption{PolyMapper results for (a) Chicago and (b) Sunnyvale. Results for Boston are shown in Fig.~\ref{fig:demo_boston}.}
    \label{fig:other_demos}
    \vspace{-1.5em}
\end{figure}

\subsection{Comparison on PolyMapper Dataset}
\label{sec:res_ours}

\begin{table*}[!ht]
    \centering
	\caption[Evaluation on the PolyMapper dataset: Buildings]{Evaluation on the PolyMapper dataset: Buildings}
	\label{tab:eval_ours_building}
	\begin{tabular}{r|r r r r r r|r r r r r r}
	\hline
	\textbf{Method} & AP & AP$_{\text{50}}$ & AP$_{\text{75}}$ & AP$_{\text{S}}$ & AP$_{\text{M}}$ & AP$_{\text{L}}$ & AR & AR$_{\text{50}}$ & AR$_{\text{75}}$ & AR$_{\text{S}}$ & AR$_{\text{M}}$ & AR$_{\text{L}}$ \\ \hline
	Mask R-CNN\cite{he2017mask,crowdAIMappingChallengeBaseline2018} & 42.0 & 70.5 & \textbf{46.7} & 24.3 & \textbf{55.5} & \textbf{49.9} & 46.6 & 71.7 & 53.6 & 27.6 & 61.1 & \textbf{60.4} \\
	PANet\cite{liu2018path} & 42.1 & 71.7 & 46.3 & 25.5 & 54.5 & 47.9 & 47.0 & 72.5 & 54.1 & 29.1 & 60.4 & 57.0 \\ \hline
	PolyMapper & \textbf{44.7} & \textbf{80.5} & 46.3 & \textbf{31.5} & 54.0 & 40.5 & \textbf{52.8} & \textbf{84.6} & \textbf{58.0} & \textbf{39.6} & \textbf{62.7} & 60.3 \\ \hline
    \end{tabular}
\end{table*}

\begin{table*}[!ht]
    \centering
	\caption[Evaluation on the PolyMapper dataset: Roads]{Evaluation on the PolyMapper dataset: Roads}
	\label{tab:eval_ours_road}
	\begin{tabular}{r|r r|r r r|r r r}
	\hline
	\textbf{Method} & SP$_{\pm\text{5\%}}$ & SP$_{\pm\text{10\%}}$ & AP$_{\text{85}}$ & AP$_{\text{90}}$ & AP$_{\text{95}}$ & AR$_{\text{85}}$ & AR$_{\text{90}}$ & AR$_{\text{95}}$ \\ \hline
	DeepRoadMapper~\cite{deeproadmapper} & 48.6 & 61.6 & 74.3 & 61.8 & 47.8 & 75.9 & 63.9 & 49.4 \\
	RoadTracer~\cite{bastani2018roadtracer} & 65.7 & 77.7 & 82.8 & 75.4 & 60.2 & 85.5 & 78.6 & 66.2 \\ \hline
	PolyMapper & \textbf{72.8} & \textbf{85.3} & \textbf{92.4} & \textbf{86.5} & \textbf{73.7} & \textbf{92.4} & \textbf{86.3} & \textbf{72.6} \\ \hline
    \end{tabular}
    \vspace{-0.5em}
\end{table*}

We are not aware of any publicly available dataset used by state-of-the-art methods that contains both annotations of building footprints and road networks for aerial imagery. Thus we created our own dataset \emph{following the same procedure} used to obtain the crowdAI~\cite{crowdAIdataset} and RoadTracer~\cite{bastani2018roadtracer} datasets. This new dataset contains building footprints and road networks from OSM~\cite{haklay2008,haklay2010,girres2010} and aerial images from Google Maps. We collect the dataset of the three US cities Boston, Chicago, and Sunnyvale. We did not choose European cities in this work because many buildings typically share the same roof and polygonal instance segmentation is thus ill-defined (i.e. a single building in the aerial image is often split into multiple instance annotations). As for Asian cities, they usually have a lot of missing annotations in OSM. 
Our new PolyMapper dataset contains $\sim$400,000 images and each patch is of size 300$\times$300 pixels and shows zoom level 19 (scale $\sim$22.57m per pixel) in Google Maps, covering 466.587$\text{km}^\text{2}$ with $\sim$3,000,000 building annotations and 8905.3km of road annotations.

Unlike RoadTracer~\cite{bastani2018roadtracer} that trains its model on 25 cities and tests on 15 different cities, we train our method and baselines on each city of the new PolyMapper dataset separately. Testing of models is done on different areas of a city (same strategy as \cite{deeproadmapper}) and a weighted average is computed across cities. Quantitative results are shown in Tab.~\ref{tab:eval_ours_building} and~\ref{tab:eval_ours_road}. We also visualize the final map extraction results for some test regions in Fig.~\ref{fig:demo_boston},~\ref{fig:demo_chicago} and~\ref{fig:demo_sunnyvale}. For more details about the statistics of the new dataset and experiments, please refer to the supplementary material.

For roads (see Tab.~\ref{tab:eval_ours_road}), PolyMapper outperforms both, DeepRoadMapper~\cite{deeproadmapper} and RoadTracer~\cite{bastani2018roadtracer} consistently across all measures (averaged across Boston, Chicago, and Sunnyvale). As for polygon building footprints extraction (see Tab.~\ref{tab:eval_ours_building}), PolyMapper performs on par with the pixel-wise instance segmentation approaches Mask R-CNN~\cite{he2017mask} and PANet~\cite{liu2018path}, but for average precision and recall, PolyMapper still outperforms them.

\section{Conclusion}
\label{sec:conclusion}

We have proposed a novel approach that is able to directly extract topological map from city overhead imagery with a CNN-RNN architecture. We also propose a novel reformulation method that can sequentialize a graph structure as closed polygons to unify the shapes of different types of objects. Our empirical results on a variety of datasets demonstrate high-level of performance for delineating building footprints and road networks using raw aerial images as input. Overall, PolyMapper performs better or on par compared to state-of-the-art methods that are custom-tailored to either building or road networks extraction in pixel level. A favorable property of PolyMapper is that it produces topological structures instead of conventional per-pixel masks, which are much closer to the ones of real online map services, and are more natural and less redundant. We view our framework as a starting point for a new research direction that directly learns high-level, geometrical shape priors from raw input data through deep neural networks to predict vectorized object representations.

\newpage

{
\small
\bibliographystyle{ieee_fullname}
\bibliography{egbib}
}

\end{document}